\documentclass{article}

     \usepackage[preprint]{neurips_2023}

\usepackage[utf8]{inputenc} 
\usepackage[T1]{fontenc}    
\usepackage{hyperref}       
\usepackage{url}            
\usepackage{booktabs}       
\usepackage{amsfonts}       
\usepackage{nicefrac}       
\usepackage{microtype}      
\usepackage{xcolor}         

\usepackage{amssymb}
\usepackage{amsthm}
\usepackage{amsfonts}
\usepackage{dsfont}
\usepackage{mathtools}
\allowdisplaybreaks

\usepackage{diagbox}
\usepackage{multirow}
\usepackage{graphicx}
\usepackage{epsfig}
\usepackage{float}
\usepackage{wrapfig}
\usepackage{subcaption}
\usepackage{algorithm}
\usepackage{algcompatible}

\usepackage{mdframed}
\usepackage{xcolor}
\newcommand{\highlight}[1]{ {\color{red}{#1}} }

\usepackage{appendix}
\usepackage{comment}
\usepackage{enumitem}
\usepackage{pdfpages}

\newtheorem{theorem}{Theorem}{\itshape}{\rmfamily}

{\itshape}{\rmfamily}

{\itshape}{\rmfamily} 	

\newtheorem{proposition}[theorem]{Proposition}{\itshape}{\rmfamily}

{\itshape}{\rmfamily}

{\itshape}{\rmfamily}

{\itshape}{\rmfamily}
{\itshape}{\rmfamily}

\newtheorem*{recap}{}{\itshape}{\rmfamily}

\newcommand{\squishlist}{
	\begin{list}{$\bullet$}
	{ 	\setlength{\itemsep}{0pt}      \setlength{\parsep}{1pt}
		\setlength{\topsep}{3pt}       \setlength{\partopsep}{0pt}
		\setlength{\leftmargin}{1.5em} \setlength{\labelwidth}{1em}
		\setlength{\labelsep}{0.5em} } }

\newcommand{\squishlisttwo}{
	\begin{list}
		{ \setlength{\itemsep}{0pt}    \setlength{\parsep}{0pt}
			\setlength{\topsep}{0pt}     \setlength{\partopsep}{0pt}
			\setlength{\leftmargin}{2em} \setlength{\labelwidth}{1.5em}
			\setlength{\labelsep}{0.5em} } }
	
\newcommand{\squishend}{\end{list}}

\newcommand{\eq}[2][]{
	\ifx\empty#1\empty
	\begin{equation*} #2 \end{equation*}
	\else
	\begin{equation}\label{#1} #2 \end{equation}
	\fi
}
\newcommand{\eqm}[2][]{
	\ifx\empty#1\empty
	\begin{equation*}\begin{split} #2 \end{split}\end{equation*}
	\else
	\begin{equation}\label{#1}\begin{split} #2 \end{split}\end{equation}
	\fi
}
\newcommand{\eqa}[2][]{
	\ifx\empty#1\empty
	\begin{align*} #2 \end{align*}
	\else
	\begin{align}\label{#1} #2 \end{align}
	\fi
}

\newcommand{\define}[0]{\doteq}

\newcommand{\pde}[2]{\frac{\partial #1}{\partial #2}}

\newcommand{\pdbyd}[1]{\frac{\partial}{\partial #1}}

\newcommand{\ra}[0]{\rightarrow}

\newcommand{\Op}[2]{#1 \Big[ #2 \Big]}
\renewcommand{\P}{\mathbf{P}}
\renewcommand{\Pr}[2][]{\P_{#1}[#2]}

\newcommand{\E}[1][]{\ifx\empty#1\empty \mathbf{E} \else \operatorname*{\mathbf{E}}_{\substack{#1}} \fi}
\newcommand{\Exp}[2][]{\Op{\E[#1]~}{#2}}

\newcommand{\cov}{\mathbf{cov}}
\newcommand{\Cov}[2][]{\ifx\empty#1\empty 
	\Op{\cov}{#2}
	\else
	\Op{\underset{\scriptscriptstyle #1}{\cov}}{#2}
	\fi}

\newcommand{\indicator}[1]{\mathds{1}[ #1 ]}

\renewcommand{\*}{\cdot}
\newcommand{\vect}[1]{\boldsymbol{#1}}

\renewcommand{\a}{ {\vect{a}} }

\newcommand{\w}{{\vect{w}}}
\newcommand{\wbar}{\w^+}

\newcommand{\M}{\mathcal{M}}

\newcommand{\seq}[1]{(#1)}
\newcommand{\X}[1][]{\ifx\empty#1\empty X \else X^{\scriptscriptstyle (#1)} \fi}
\newcommand{\Y}[1][]{\ifx\empty#1\empty Y \else Y^{\scriptscriptstyle (#1)} \fi}
\newcommand{\Z}[1][]{\ifx\empty#1\empty Z \else Z^{\scriptscriptstyle (#1)} \fi}
\renewcommand{\L}[1][]{\ifx\empty#1\empty L \else L^{\scriptscriptstyle (#1)} \fi}
\newcommand{\Yp}[2][]{\ifx\empty#1\empty Y_{<#2} \else Y^{\scriptscriptstyle (#1)}_{<#2} \fi}
\newcommand{\Zp}[2][]{\ifx\empty#1\empty Z_{<#2} \else Z^{\scriptscriptstyle (#1)}_{<#2} \fi}

\newcommand{\x}[1][]{\ifx\empty#1\empty \vect{x} \else \vect{x}^{\scriptscriptstyle (#1)} \fi}
\newcommand{\y}[1][]{\ifx\empty#1\empty \vect{y} \else \vect{y}^{\scriptscriptstyle (#1)} \fi}
\newcommand{\z}[1][]{\ifx\empty#1\empty \vect{z} \else \vect{z}^{\scriptscriptstyle (#1)} \fi}
\newcommand{\h}[1][]{\ifx\empty#1\empty \vect{h} \else \vect{h}^{\scriptscriptstyle (#1)} \fi}
\newcommand{\xt}[2][]{\ifx\empty#1\empty x_{\scriptscriptstyle #2} \else x^{\scriptscriptstyle (#1)}_{\scriptscriptstyle #2} \fi}
\newcommand{\yt}[2][]{\ifx\empty#1\empty y_{\scriptscriptstyle #2} \else y^{\scriptscriptstyle (#1)}_{\scriptscriptstyle #2} \fi}
\newcommand{\zt}[2][]{\ifx\empty#1\empty z_{\scriptscriptstyle #2} \else z^{\scriptscriptstyle (#1)}_{\scriptscriptstyle #2} \fi}
\newcommand{\at}[2][]{\ifx\empty#1\empty h_{\scriptscriptstyle #2} \else h^{\scriptscriptstyle (#1)}_{\scriptscriptstyle #2} \fi}

\newcommand{\yp}[2][]{\y[#1]_{<#2}}

\newcommand{\bos}{\texttt{bos}}
\newcommand{\eos}{\texttt{eos}}
\renewcommand{\i}{^{\scriptscriptstyle (i)}}

\newcommand{\B}{{ \mathcal{B} }}
\newcommand{\Bopt}[1][]{{\ifx\empty#1\empty \B \else \B^{#1} \fi}}

\newcommand{\Lagr}{{ \mathcal{L} }}
\newcommand{\mul}{{ \vect{\lambda} }}

\newcommand{\gammaEPI}{{ \gamma_\text{~epi~} }}

\newcommand{\qcov}[2]{\cov_{#1} \Big[ Q, \pde{Q}{w_{#2}} \Big]}

\newcommand{\Pw}{\P_{\w}}
\newcommand{\Qw}{Q({\w})}
\newcommand{\Ptrue}{\P_{\text{true}}}
\newcommand{\supp}{\mathrm{supp}}

\newcommand{\q}{\vect{q}}
\newcommand{\p}{\vect{p}}
\newcommand{\Hmap}{\vect{y}_{\texttt{MAP}}}
\newcommand{\ymap}{\vect{y}_{\texttt{MAP}}}
\newcommand{\ygreedy}{\y_{\text{greedy}}}
\newcommand{\Hpi}{A}
\newcommand{\Jmle}{J_{\texttt{MLE-Q}}}
\newcommand{\Jmabe}{J_{\text{MABE}}}
\newcommand{\Pdual}{p^\text{dual}}

\title{Utility-Probability Duality of Neural Networks}

\author{%
	Huang Bojun~\thanks{Correspondence to: Huang Bojun < \texttt{bojhuang@gmail.com} >} \\
	Rakuten Institute of Technology, \\
	Rakuten Group, Inc. \\
	\texttt{bojhuang@gmail.com}
	\And
	Fei Yuan \\
	Shanghai Artificial Intelligence \\ Laboratory \\
	\texttt{yuanfei@pjlab.org.cn}
}

\begin{document}

\maketitle

\begin{abstract}
It is typically thought that supervised training of modern neural networks is a process of fitting the groudtruth probabilities. However, many counter-intuitive observations in language generation tasks let one wonder if this canonical probabilistic explanation can really account for the observed empirical success.  
To resolve this issue, we propose an alternative \emph{utility-based explanation} to the standard supervised learning procedure in deep learning. The basic idea is to interpret the learned neural network not as a probability model but as an \emph{ordinal utility} function~\cite{1990:utility} that encodes the preference revealed in training data. We developed a theory based on this utility-based interpretation, in which the theoretical expectations and empirical observations are better reconciled.

\end{abstract}

\section{Introduction}
In this paper we challenge, and fix, a standard explanation of deep learning. The 
prevailing mindset nowadays is to \emph{interpret} a neural network $f(\w)$ as a parametric model of the conditional probability distribution $\Pw[Y=\y|X=\x]$, where $X$ is an \textbf{expected input} of the task under concern (e.g. an image/sentence/speech audio), and $Y$ is an \textbf{expected output} given $X$ (e.g. a class label, a score, or a structured object such as a sentence or an action plan) which is assumed to follow a groundtruth distribution $\Ptrue$. Training of the neural network $f(\w)$ is then \emph{thought} to be the process of approximating $\Ptrue$ with $\Pw$. Indeed, in \cite{2016:dl}, the deep learning textbook writes (p.138): ``\emph{most supervised learning algorithms in this book are based on estimating a probability distribution $p(y|x)$}''.

This probability interpretation of neural networks supports two natural ways to use the learned probability model $\Pw$ at inference time. The first way is to choose the most likely output in $\P_\w$: 
\eq[map]{
	\ymap \define \arg\max_{\y}~ \Pr[\w]{Y=\y|X=\x}
}
where MAP stands for \emph{maximum a-posteriori probability}. 
When $\Pw = \Ptrue$, $\ymap$ is a provably optimal output in many common scenarios~\cite{2006:bishop}; 
see Appendix \ref{sec:map} for a rigorous analysis.

Another sensible decision rule is to sample the output from the distribution $\Pw$, which makes the \textbf{actual output} a random variable, denoted by $\Hpi$ here: 
\eq[sampling]{
	\Hpi \sim \Pr[\w]{Y=\cdot|X=\x}
}
When $\Pw = \Ptrue$, the stochastic output $A$ is not necessarily optimal, 
but is necessarily a good output as long as the expected output $Y$ is the output of a good decision policy (because $A$ is identically distributed with $Y$; see Appendix \ref{sec:sampling} for more elaboration on the soundness of the sampling rule).

The two decision rules \eqref{map} and \eqref{sampling} underlie a long tradition in the ML community that reduces the problem of \emph{learning to make decisions} to a probability estimation problem~\footnote{
	The tradition was there before the deep learning era. In another classic ML book, \citet{2006:bishop} writes ``\emph{determination of $p(\x,t)$ ... forms the subject of much of this book}'' (p.38).
}: If we could estimate $\Ptrue$ perfectly, our decision would be guaranteed good. In reality, the approximation of $\Ptrue$ with $\Pw$ always comes with errors, but the correspondence in the limit between decision making and probability estimation still gives the reasonable expectation that the closer the probability estimation is, the better the induced decision -- by the two decision rules \eqref{map} and \eqref{sampling} -- would be. 

However, it is known that 
many neural networks with excellent decision quality are actually poorly calibrated in terms of probability estimation~\cite{2017:calibration, 2021:calibration}. In fact, \citet{2017:calibration} reported that for some popular NN architectures, more powerful models (in terms of classification quality) tend to be worse calibrated in terms of how $\Pw$ matches $\Ptrue$.

More importantly, recent empirical studies in NLP found that for a variety of language generation tasks, both the MAP rule $A=\ymap$ and the sampling rule $A\sim \Pw$ lead to very bad performance in terms of text/decision quality~\cite{2019:stahlberg, 2019:cohen, 2020:mbr, 2019:sampling}; this is the case even for extensively-trained models with state-of-the-art architectures. On the other hand, greedy or near-greedy outputs, as a kind of ``economic yet sub-optimal'' choices from the probabilistic perspective, turn out to work significantly better, and is often the \emph{only} solution that is known to work satisfactorily, not in cost, but in quality~\cite{2016:gnmt}. 
These paradoxical observations form an explainability issue that challenges the probabilistic rationale behind the empirical success in related domains (Section \ref{sec:paradox}).

To resolve this issue, we propose an alternative \emph{value-based explanation} to the standard supervised learning procedure in deep learning. The basic idea is to interpret the learned neural network not as a probability model but as an \emph{ordinal utility} function~\cite{1990:utility, 2018:measure_utility} that encodes human preference revealed in training data. We develop a theory based on this value-based interpretation, in which the theoretical expectations and empirical observations are better reconciled.

Specifically, in Section \ref{sec:duality} we point out that a softmax-normalized neural network model also comes with an un-normalized sub-model for the logits, and that this logit sub-model is the actual functioning part of the overall model at inference/decision time. As a result, the standard MLE training process for softmax probability model can be equivalently seen as a certain learning dynamic for the un-normalized sub-model. Now suppose we could \emph{directly} explain why the sub-models trained with this particular learning dynamic \emph{will} support good greedy decisions -- in a way that the explanation does not resort to probabilistic semantics of the (sub-)model -- then the probability-based interpretation would become unnecessary and can be bypassed. What is bypassed together is the contradiction between the probabilistic interpretation and experimental observations.

In Section \ref{sec:q-learning}, we indeed provide such a non-probabilistic explanation. Without the probabilistic semantic, the ``logit sub-model'' is re-interpreted as just a Q-function, and we show that the ``MLE-equivalent learning dynamic'' of this Q-function is a perturbed variant of a particular supervised Q-learning algorithm family (called \emph{MABE}). We mathematically prove that the unperturbed variant of this family is indeed training the Q-function toward an \emph{optimal} utility function that gives optimal output under greedy decision. Then we experimentally intervene the perturbation term, and show that the perturbation (which makes the ``MLE-equivalent variant'' different from the unperturbed variant) has little impact on the learning dynamic empirically. 

Moreover, in Section \ref{sec:calibration} we derive an equation from this utility-based theory which allows us to transform the learned Q-values back to estimations of $\Ptrue$, thus bringing back the probabilistic semantic. However, the utility-based probability estimation, called \emph{dual probability} in the paper, encodes a different probability space from the canonical softmax probabilities. Intriguingly, running probability-based decision rules \eqref{map} and \eqref{sampling} based on the dual probability leads to \emph{dramatically} better performance in all tasks we examined (e.g. +14.6 BLEU for sampling and +17.3 for MAP in WMT'14 en2de translation), and it also gives more reasonable probability predictions. This result implies that the standard supervised learning procedure in deep learning -- as a utility-learning procedure now -- may indeed correspond to a dual process of probability learning, \emph{but}, the probability space learned from this dual process may not be best represented by the softmax probabilities as usually perceived. 

Overall, all the evidences above collectively revealed a phenomenon of \emph{utility-probability duality}, that a neural network is perhaps both a utility function and a probability function in many deep learning settings (Section \ref{sec:discussion}).

\section{The Paradox}
\label{sec:paradox}

It is relatively well known that the probabilities predicted by many deep neural networks (that well support decision making in practice) do not match the true probabilities very well~\cite{2017:calibration}. 
But this observation 
alone does not necessarily contradict with the probabilistic rationale behind neural network learning. 
The genuine paradox 
manifests itself through a \emph{reversal} in terms of the quality between ``supposedly-rational'' and ``supposedly-irrational'' decisions from the probabilistic perspective. Such a reversal was observed in a variety of \emph{language generation} tasks, such as machine translation~\cite{2017:six_challenges}, abstractive summarization~\cite{2019:cohen}, and image captioning~\cite{2019:sampling}. 
In this work we used three such tasks for experimentation: \textbf{WMT'14 English$\ra$German (en2de)}, the most-widely used machine translation (MT) benchmark, consisting of $4.5$ millions training sentences; \textbf{WMT'17 Chinese$\ra$English (zh2en)}, another classic MT benchmark where the source and target languages are remote, consisting of $21$ millions training examples; \textbf{CNN/DailyMail}, the most-widely used benchmark for abstractive document summarization, consisting of nearly $300$ thousands of document-abstract pairs. 
%
%

\subsection{The Expectations}
In these tasks, the expected output $\y=(\yt{1},\yt{2},\dots,\yt{T})$ consists of a sequence of atomic decisions; each $\yt{t}$ is called a \emph{token} without loss of generality. In this case, a neural network $f(\w)$ is usually thought to be an \emph{auto-regressive model} that represents the token-wise conditional probabilities: $f_{[\yt{t}]}(\x,\yp{t}; \w) = \Pr[\w]{y_t|\x, \yp{t}}$, where $f_{[\yt{t}]}$ denotes the vector-component of $f$'s output that corresponds to token $\yt{t}$, and $\yp{t} \define (\yt{1},\yt{2},\dots,\yt{t-1})$ denotes the partial output up to decision step $t$.
Such a neural network encodes $\Pr[\w]{\y|\x}$ through the \emph{product rule of probability}, with 
%
\eq[prob_y]{
	\Pr[\w]{\y|\x} = \prod\nolimits_t \Pr[\w]{\yt{t}|\x,\yp{t}} = \prod\nolimits_t f_{[\yt{t}]}(\x,\yp{t}; \w)
	.}
By substituting \eqref{prob_y} into \eqref{map} and \eqref{sampling}, we \emph{can} effectively maximize or sample $\Pw$ using the neural network $f(\w)$ as we did in one-shot decision tasks. In particular, finding the MAP output (over all possible sentences of a natural language) is a shortest-path problem that can be solved by a backtracking search in realistic (yet costly) time, as demonstrated by \cite{2019:stahlberg}. 

On the other hand, a seemly sub-optimal but economic choice is to simply output a sequence $\y$ where each token $\yt{t}$ is a greedy decision that \emph{locally} maximizes the token-wise probability given the auto-regressive context:
\eq[greedy]{
	\ygreedy \define (\yt{1}\dots \yt{T}) ~~\text{where}~~	
	\yt{t} = \arg\max_{a}~ \Pr[\w]{a|\x,\yp{t}}
} 
Comparing \eqref{map} and \eqref{greedy}, we see that the MAP decision rule maximizes over the combinatorial space of all possible sentences (= the output space), while the greedy decision rule maximizes over the token space, which avoids the combinatorial search at the cost of returning outputs with lower predicted probability. It is thus expected that the MAP rule \emph{should} give higher quality outputs than the greedy rule \emph{if} the model-predicted probability is a good indicator of the true likelihood $\Ptrue$. 

In practice, \emph{beam search} is a popular generalization of the greedy rule that generates near-greedy outputs by making each decision step based on not a single but a pool of auto-regressive contexts. With the capacity of the pool, a.k.a. the \emph{beam size}, being $1$, beam search degenerates exactly to the greedy rule \eqref{greedy}. With the beam size tuned toward infinity, beam search will eventually (but extremely slowly) cover the entire output space and will return the MAP output \eqref{map} in the theoretical limit. It is thus expected that the output quality of beam search should improve as beam size grows.


Moreover, the sampling rule \eqref{sampling} \emph{should} have reasonable performance \emph{if} the neural network is well modeling the probabilities of a desired output (see Appendix \ref{sec:sampling}). In other words, if sampling a probability model cannot give reasonable outputs, it must be because the model is not well modeling the true probabilities -- in that case there is no reason to expect that picking the greedy tokens would give something significantly better.   



\subsection{The Observations}
\label{sec:paradox_observations}
Surprisingly, however, in a number of language generation tasks, the greedy rule \eqref{greedy} and its close variants perform \emph{much} better than the more principled decision rules \eqref{map} and \eqref{sampling}. Figure \ref{fig:mle_sampling_wmt14en2de} illustrates the issue in WMT'14 en2de, with an experiment we designed to synthesize all the related counter-intuitive observations together in a systematic and self-contained way. In this experiment, a Transformer neural network with 60 millions parameters was trained using the standard MLE loss (see Appendix \ref{sec:wmt14en2de} for experiment setting details). From Figure \ref{fig:mle_sampling_wmt14en2de} we see that:


\begin{figure}[t] 
	\begin{minipage}{0.5\linewidth}
		\centering
		\includegraphics[width=1.0\linewidth]{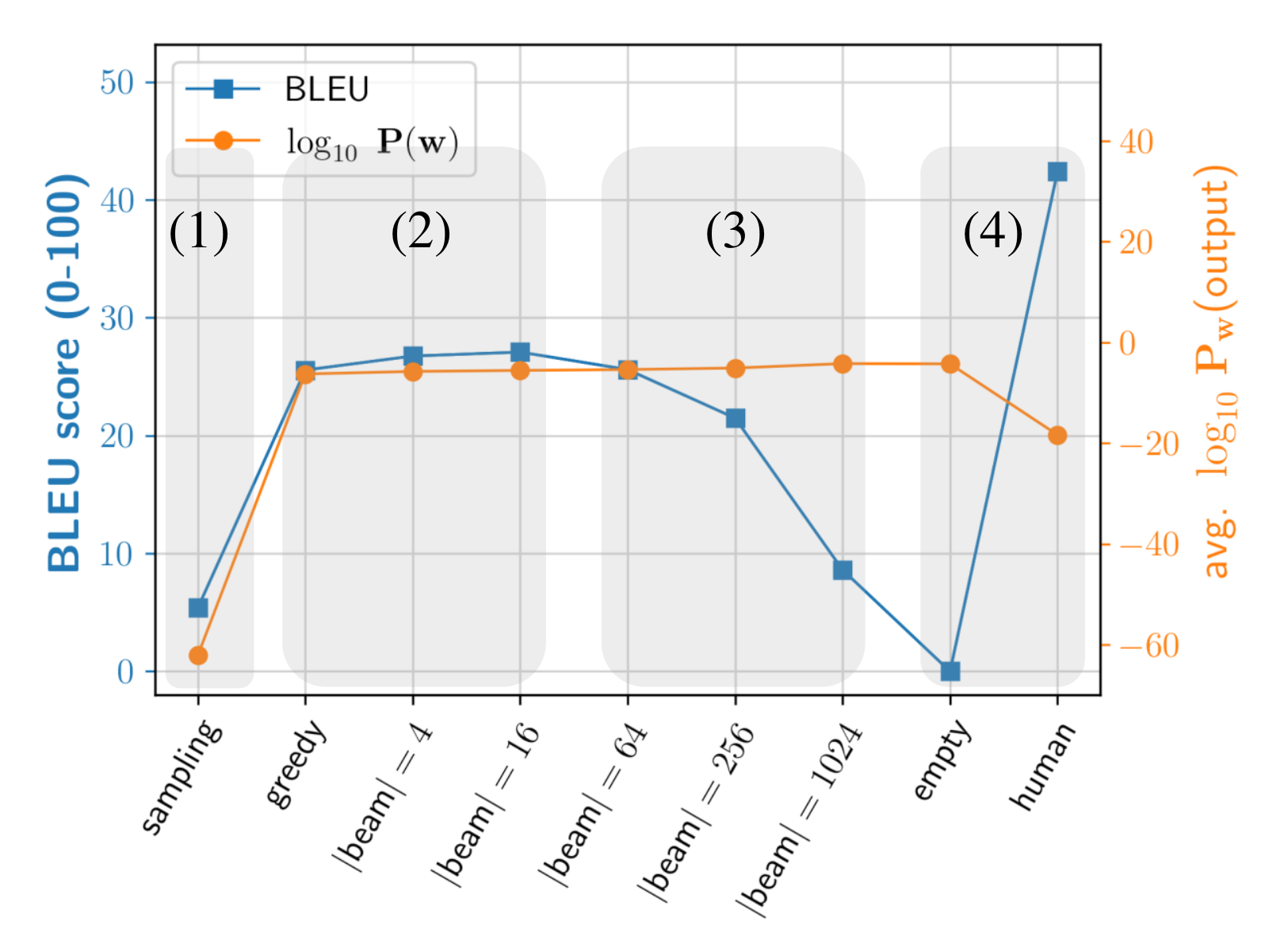}
\end{minipage}
\hfill
\begin{minipage}{0.5\linewidth}
	\centering
	\scalebox{0.95}{\begin{tabular}{|l|c|c|}
			\hline
			& BLEU 	& $\log_{10} \Pw$ \\
			\hline
			Empty output						& ~~~0.0 			& ~~\highlight{-4.3}  \\
			\hline
			Sampling rule \eqref{sampling}		& ~~~\highlight{5.4} & -62.2  \\
			\hline
			Greedy	rule \eqref{greedy}			& 25.6 				& ~~-6.3  \\
			\hline
			$|\text{beam}|=16$ (near-greedy)	& 27.1 				& ~~-5.6  \\
			\hline
			$|\text{beam}|=1024$ (approx. MAP) 	& ~~\highlight{8.6} & ~~-4.2  \\
			\hline
			MAP rule \eqref{map}				& ~$\approx$\highlight{2.1}\footnote{\citet{2019:stahlberg} reported BLEU$=2.1$ for the true MAP output on WMT'15 en2de.} & $>$~-4.2~~  \\
			\hline
			Human's output		& ~~~42.5~\footnote{See calculation method in Appendix \ref{sec:wmt14en2de}.} & \highlight{-18.4}  \\
			\hline
	\end{tabular}}
\end{minipage}
\caption{MLE models exhibit paradoxical observations in WMT'14 en2de Translation. The performance is measured by the standard BLEU metric in the domain~\cite{2002:bleu}. $\log_{10} \Pw$ gives the order-of-magnitude of the probability as predicted by the trained model. 
	\vspace{-0.1in}
}
\label{fig:mle_sampling_wmt14en2de}
\end{figure}

(1) \textbf{Sampling the learned probability model $\Pw$ gives bad outputs}. Specifically, the performance of sampled output (corresponding to ``sampling'' in Figure \ref{fig:mle_sampling_wmt14en2de}) is $5.4$ ($\pm$0.3 for $90\%$ confidence interval). As the baselines, human's score is $42.5$, and the best machine translation solution (``$|$beam$|$=16'' in Figure \ref{fig:mle_sampling_wmt14en2de}) is $27.1$. The performance of sampling the probability model is much closer to random output's ($\approx 0$), which is far from being a reasonable performance.

(2) \textbf{The greedy rule gives good outputs}, 
achieving a performance score of $25.6$. While another $1.5$ score can be obtained by relaxing the greedy pick to a few candidates each time (``$|$beam$|$=16'' in Fig.\ref{fig:mle_sampling_wmt14en2de}), the gap is rather marginal. In general, it is fair to say that greedy outputs are nearly the best, or that near-greedy outputs are the best.

(3) \textbf{Seeking to maximize the predicted probability gives bad outputs}: As we continue to increase the beam size, we indeed find outputs with higher probability according to the model (orange curve), but the actual translation quality turns out to decrease (blue curve). With beam size $=1024$, the performance has dropped to $8.6$. 
In fact, \citet{2019:stahlberg} reported that the exact MAP output has a performance score as low as $2.1$ in a similar WMT'15 task.

(4) \textbf{The learned probability model $\Pw$ significantly over-estimates some clearly bad outputs, while under-estimates, again significantly, some clearly good outputs on the other hand}. Specifically, the model predicts a probability of around $1/10^4$ for the empty translation which consists of nothing but an end-of-sequence token -- clearly such translation should never occur (and indeed the model never saw any empty translation in the training data).~\footnote{
Probability over-estimation is not limited to this particular output. It is a general trend that current probability models over-estimate many very short and meaningless outputs~\cite{2016:gnmt,2018:length_bias}.
} 
On the other hand, the model assigns \emph{lower} probability to the best-performing outputs (e.g. $\Pw[\ygreedy]\approx 1/10^6$ in Figure \ref{fig:mle_sampling_wmt14en2de}), and moreover, assigns \emph{much lower} probability to the true expected outputs provided by human (with average predicted probability as low as $1/10^{18}$).

The above observations are not limited to the particular task as demonstrated. See Appendix \ref{sec:wmt17zh2en} and \ref{sec:cnndm} for similar results in WMT'17 zh2en translation and in CNN-DM summarization, respectively. The pattern is the same across all tasks: Both maximizing and sampling the learned probability model perform poorly while going greedy or near-greedy with the ``local probabilities'' performed dramatically well, and the learned model systematically assigns very low probabilities to desired outputs while giving much higher probabilities to undesired outputs.

These observations lead to a paradox if we insist the probabilistic explanation: On one hand, we see strong reasons to reject the greedy rule -- for three tokens $A$, $B$, $C$, saying that ``\emph{$AC$ is more likely than $BC$ because $P(A)>P(B)$}'' (which is what the greedy rule is suggesting!) violates basic principles of probability theory. On the other hand, we do observe that the greedy rule works very well in reality, much better than decisions based on $P(AC)$ and $P(BC)$.

\subsection{Existing Explanations}

The counter-intuitive observations above make one naturally wonder if the learned neural networks are really supporting decision making \emph{through} good probability modeling. Indeed, aspects of this problem have been called ``\emph{beam search curse}''~\cite{2018:beam_search_curse}, ``\emph{beam search bless}''~\cite{2020:uid}, or ``\emph{neural text degeneration}''~\cite{2019:sampling} -- these names may have suggested how paradoxical the community are feeling about the problem. In the following we briefly mention some existing explanations in the literature. See Appendix \ref{sec:existing_explanations} for an extended discussion of related works. 
%
%

\citet{2020:mbr} defended the probabilistic interpretation, arguing that the MAP output is not a good decision rule at all for the selected task. We however argue that tasks like translation fall in the category that MAP is provably optimal \emph{if} the probability estimation is accurate (see Appendix \ref{sec:map}), so inadequacy of MAP output can only be \emph{caused} by inaccuracy of probability modeling.

Many works \cite{2016:ranzato, 2019:bridging, 2016:gnmt, 2019:stahlberg, 2019:cohen, 2019:sampling} seek to find out why the learned model deviates from the groundtruth distribution. Factors such as \emph{exposure bias}~\cite{2020:exposure}, \emph{length bias}~\cite{2016:gnmt}, abnormal probability fluctuations~\cite{2019:cohen}, and long-tail errors~\cite{2019:sampling}, are identified, with many heuristic methods proposed to avoid the identified failure patterns.
This line of works however did not explain why (near-)greedy decisions \emph{based on a model with so many issues} can somehow lead to good empirical result. Note that the heuristics proposed in these works themselves have effectively made the resulted solution deviating from the probability principles. It is still unclear why we \emph{have to} violate well-established probability principles, either in the form of greedy decision or by some sort of heuristic rules, to obtain competitive performance from the learned ``probability models''.



\citet{2020:uid} recently proposed to explain the effectiveness of greedy output via a ``uniform information density (UID)'' hypothesis in cognitive science. However, the precise mathematical expression of the UID hypothesis itself is subject to different interpretations~\cite{2021:uid}. In contrast, in this paper we propose a mathematically accurate and non-probabilistic explanation.


\section{A Duality View to MLE Training}
\label{sec:duality}

We lay out a conceptual framework in this section which aims at resolving the paradox as illustrated in Section \ref{sec:paradox} through a shift of mindset. Consider the standard MLE training process for neural networks: We collect a data set $\{\x[i], \y[i]\}_{i=1..n}$ 
with $\y[i] \sim \Pr[\text{true}]{Y|X=\x[i]}$, then we update the model parameters $\w$ toward the ones that maximize the (log-)probability of the data in $\Pw$:
\eqm[mle]{
\w_{\texttt{MLE}} &= \arg\max_{\w}~ \log \Pr[\w]{\{\y[1..n]\} | \{\x[1..n]\}} 
= \arg\max_{\w}~ \sum_{i=1}^n \sum_{t=1}^{T_i} \log f_{[\yt[i]{t}]}(\x[i],\yp[i]{t}; \w)
}
where the neural network $f$ models a softmax distribution
\eq[softmax]{
f_{[\yt{t}]}(\x,\yp{t};\w) 
= e^{Q_{[\yt{t}]}(\x,\yp{t};\w)} / \sum\nolimits_a e^{Q_{[a]}(\x,\yp{t};\w)}
}
over the so-called \emph{logit} vector $Q(\x,\yp{t};\w)$. The training of $f(\w)$ follows a learning dynamic driven by the gradient of the MLE objective \eqref{mle}.
\footnote{
See Appendix \ref{sec:mle} for how \eqref{mle} corresponds \emph{exactly} to the NLP practice in real world.
}  
This gradient can be conveniently computed by substituting \eqref{softmax} into \eqref{mle}, yielding
\begin{align}
\nabla_{\w} \log \Pr[\w]{\yt{t}|\x,\yp{t}} 
=~& \nabla_{\w} Q_{[\yt{t}]}(\w) - \nabla_{\w} \text{LogSumExp}\Big( Q(\w) \Big) \nonumber\\
=~& \nabla_{\w} Q_{[\yt{t}]}(\w) - \sum\nolimits_a ~ f_{[a]}(\w) ~ \nabla_{\w} Q_{[a]}(\w) \label{grad_lprob}
\end{align}
where $\text{LogSumExp}(Q) \define \log \sum_a \exp(Q_{[a]})$, and we have omitted $(\x,\yp{t})$ in $Q$'s and $f$'s argument for brevity. 
\eqref{grad_lprob} is a simple fact known by many, and is often utilized for the purpose of computing the gradient of the log likelihood function.

We however argue that one can also understand 
\eqref{grad_lprob} in the opposite direction. Instead of viewing \eqref{grad_lprob} as a method to implement a learning dynamic of $\Pw$ through manipulating $\Qw$, we can reversely interpret it  
as a method to implement a learning dynamic of $\Qw$ through manipulating $\Pw$. In this alternative perspective, we iterate the function $Q$
\footnote{
In the new perspective we shall perhaps not call $Q$ the ``logits'' any more, a name that itself is suggesting that $Q$ is nothing but the logarithm of something else.
} 
for its own sake, in the particular way as prescribed by the right-hand side of \eqref{grad_lprob}.
The left-hand side of \eqref{grad_lprob}  
-- as well as its connection to the MLE objective \eqref{mle} -- is merely a human-imposed explanation about this Q-oriented learning dynamic. More generally, the iteration of $\Pw$ and the iteration of $\Qw$ can be considered \emph{dual process} to each other that are taking place in parallel, in a learning dynamic conventionally named ``MLE training''.


While in principle one is free to choose either the P-iteration view or the Q-iteration view, 
the former (i.e. the probabilistic interpretation) will induce many conflicts with the empirical observations as shown in Section \ref{sec:paradox}. For this reason, we propose to explain the empirical behaviors of softmax-normalized neural networks from the Q-iteration perspective, in which what the neural network is expected to output are not probabilities, but are just the un-normalized Q-values. The Q-function is trained with the RHS of \eqref{grad_lprob} being the update rule. At decision time, the well-performed greedy-to-P rule \eqref{greedy} (which requires probabilistic interpretation) can be re-interpreted as a greedy-to-Q rule:
\eq[greedyQ]{
	\ygreedy = (\yt{1 \dots T}) ~\text{,}~~	
	\yt{t} = \arg\max_{a}~ Q_{[a]}(\x,\yp{t}; \w)
} 
because the softmax transformation is order preserving.

Note that in above we are \emph{not} proposing a new algorithm, but were only rephrasing the standard training and inference procedures in existing practice (i.e. \eqref{mle} and \eqref{greedy}) from another point of view. As the probabilistic semantic is entirely discarded, all the probability-based assertions about the softmax outputs become unexpected, thus the weak performance of probability-based decision rules and the unreasonable probability predictions are not paradoxical any more under the Q-iteration perspective. The only thing that needs to be explained is \emph{why} the particular Q-iteration procedure \eqref{grad_lprob} \emph{can} learn a good Q-function for greedy usage, which would be the main topic of the next section.

Before turning to our account to the above question, we first remark that ``\emph{learning unnormalized Q-functions in support of greedy decision making}''
is not a random problem we posed here just for fitting a particular experimental result, but is a classic research topic that has been extensively studied in reinforcement learning (RL)~\cite{1989:qlearning, 2015:dqn, 2018:RL}. 
In RL literature, such a Q-function is also called an \emph{action-value} function, or just \emph{value function} for short. Value functions support decision making by assigning preferential scores to options so that optimal ones can be identified, locally and greedily, without checking the long-term consequence of the local decision. A value function is called an \textbf{optimal Q-function} if the induced greedy decision policy \eqref{greedyQ} gives optimal outputs. 

However, in existing RL literature, the concept of value has been mostly limited to a particular type of \emph{cardinal utility} which corresponds to the expected reward of a specific policy, and in RL the optimal Q-function is typically learned via a \emph{Bellman value iteration} procedure. The manipulations on the Q-function in the ``MLE dynamic'' \eqref{grad_lprob} is clearly a very different procedure. In fact, different from the typical RL setting where the learning is driven by a reward signal, the dual process \eqref{grad_lprob} of MLE optimization relies on demonstrative samples of the expected output -- in other words, it is an \emph{imitation learning} procedure of Q-learning. The Q-function thus learned represents an ordinal utility function that does not necessarily correspond to some pre-defined reward; see Appendix \ref{sec:value} for more comparison and contrast between the utility concept and the value concept in RL. Existing RL or imitation learning literature thus cannot fully explain the utility-based rationale behind this uncommon (but empirically effectively) procedure:
If the standard supervised MLE training of deep neural networks is actually learning Q-functions, what is the ``target'' of this Q-learning dynamic? Is the learning steered toward an \emph{optimal} Q-function?  Can we explain this procedure, which works well when (and to large extent, only when) coupled with the greedy decision rule, without resorting back to the probability interpretation? We seek to address these questions in the next section.

\section{MLE Training as a Perturbed Dynamic of Optimal Utility Learning}
\label{sec:q-learning}

Our general goal is to interpret the SGD dynamic of MLE training as an SGD process of \emph{some} objective function of $Q$. There is however a technical obstacle: If we look at \eqref{grad_lprob}, the gradient operator $\nabla_{\w}$ cannot be re-arranged to the head because of the $f_{[a]}(\w)$ term. As a result, it is not immediately clear that \eqref{grad_lprob} is computing a gradient of anything other than the log-likehood (which is the interpretation we wanted to bypass here). 

Nonetheless, let us temporarily do a ``wrong'' algebraic manipulation by moving $\nabla_{\w}$ to the head of \eqref{grad_lprob} anyway, making it ``approximately'' the gradient of the following objective function:
\vspace{-0.1in} 
\eqa[J_mabe]{
\Jmabe(\w,\x,\y) 
&	\define  
	\sum_{t=1}^{T}~ \Big(~ 
		Q(\yt{t}|\x,\yp{t};\w) - \sum\nolimits_{a} f(a|\x,\yp{t};\w) Q(a|\x,\yp{t};\w) 
	~\Big)
\nonumber \\
&= 	
\sum_{t=1}^{T}~ \Big(~ 
Q(\yt{t}|\x,\yp{t};\w) - \Exp[\Hpi_t \sim \P_{\w}]{ Q(\Hpi_t|\x,\yp{t};\w) } 
~\Big)
}
Intuitively, $\Jmabe$ measures the \emph{advantage} of outputting the expected token $y_t$ over a stochastic output $\Hpi_t$ that follows the softmax distribution induced by $Q$, where the advantage is the difference of expected utilities between the two decisions (as predicted by $Q$, under context $\x,\yp{t}$, for each step $t$). The subscript MABE stands for Maximum Advantage over Boltzmann Exploration.

From \eqref{grad_lprob} to \eqref{J_mabe}, we have (naively) understood the log-likelihood gradient $\nabla_{\w} \log \Pw$ as an approximation of $\nabla_{\w} \Jmabe$, with the impact of $\partial \w$ over $f(\w)$ being ignored. In this sense, the MLE optimization can be seen as a \emph{biased} SGD dynamic of $\Jmabe$ optimization. 

Interestingly, the log-likelihood gradient is not arbitrarily biased, but there is a \emph{precise} connection between the gradients of the two functions (see the proof in Appendix \ref{sec:proof_mle}):

\begin{proposition}
\label{thm:grad_greedyq}
Given input $\x$, output $\y=\seq{\yt{1} \dots \yt{T}}$, and parametric model $Q(\w)$, 
we have
\eqm{
	\pde{\log \Pr[\w]{\y|\x}}{w_j}  =  
	\pde{\Jmabe}{w_j} (\w,\x,\y) ~+~ \sum_{t=1}^T \qcov{t}{j}   \quad\quad,~~ \forall j
}
where~ $\qcov{t}{j} \define \Cov[\Hpi_t \sim P_t]{Q_t(\Hpi_t) ~,~ \pde{Q_t}{w_j}(\Hpi_t)}$
\eqa[cov]{
	=~&	\sum_a P_t(a) \cdot \Big(~ Q_t(a) - \sum_{b} P_t(b) Q_t(b) ~\Big) 
	\cdot \Big(~ \pde{Q_t}{w_j}(a) - \sum_{b} P_t(b) \pde{Q_t}{w_j}(b) ~\Big) 
}
\vspace{-0.1in}
\squishlist
\item $\P_{\w}$ is the softmax distribution induced by $Q(\w)$ as prescribed by \eqref{prob_y} and \eqref{softmax},
\item $P_t(a)$ denotes $\Pr[\w]{a|\x,\yp{t}}$, the probability that the token at step $t$ is $a$, according to $\Pw$,
\item $Q_t(a)$ denotes $Q(a|\x,\yp{t};\w)$, the utility of outputting token $a$ at step $t$, according to $Q(\w)$,
\item $\pde{Q_t}{w_j}(a)$ denote $\pde{Q}{w_j}(a|\x,\yp{t};\w)$, the partial derivative of $Q(a|\x,\yp{t};\w)$ at step $t$,
\squishend
\end{proposition}

Intuitively, the $\cov_t$ term in \eqref{cov} is the covariance between $Q(A_t)$ and its derivative when $A_t$ follows the Boltzmann exploration policy $\Pw$. Proposition \ref{thm:grad_greedyq} asserts that the gradient of the probability-learning objective \eqref{mle} differs from the gradient of the utility-learning objective \eqref{J_mabe} by exactly this covariance (or by the cumulative covariance in sequential decision setting). 
For complex models with millions or billions of parameters, if the model output is not strongly correlated to the partial derivative of a single parameter, the covariance term identified in Proposition \ref{thm:grad_greedyq} would have limited impact on the learning progress. 
As a key result, we empirically found that this is indeed true: in all the tasks we have experimented, the perturbation from this covariance term cannot significantly affect the learning, not only for the final performance, but also throughout the entire learning dynamic. 

Specifically, consider a MABE($\lambda$) family of Q-learning algorithms as defined by Algorithm \ref{algo:mabe}. MABE(0) optimizes $\Jmabe$ based on unbiased estimator of $\nabla \Jmabe$. MABE(1) adds the covariance term to the gradient estimator, thus is equivalent to traditional MLE training. For other $\lambda$, MABE($\lambda$) does not seem to have principled interpretations, but is simply constructed by perturbing the gradient estimator with a $\lambda$ multiple of the covariance, where $\lambda$ can be either positive or negative. By tuning $\lambda$ to different values, we can control how significantly the gradient is perturbed.

\begin{algorithm}[t]
	\caption{The MABE($\lambda$) algorithm.}
	\label{algo:mabe}
	\begin{algorithmic}
		\STATE {\bfseries Input:} A supervised training data $\mathcal{D}$; a Q-model $Q(\w)$ with $|\w|=d$; perturbation coefficient $\lambda$.
		\FOR{SGD step $k = 0, 1, 2, \dots$}
		\STATE obtain a minibatch $\{\x[i], \y[i]\}_{i=1}^B$ from $\mathcal{D}$
		\STATE set $\Delta \gets \frac{1}{B} \sum_{i=1}^B \Delta\i$, where
		\STATE $\Delta\i = \nabla \Jmabe(\w,\x[i], \y[i]) + \lambda ~ \cov\i \Big|_{\w=\w^{k}}$
		and $\cov\i  = \Big< \sum_{t=1}^{T_i} \qcov{t}{j} \Big>_{j=1..d}$
		\STATE update $\w$ using $-\Delta$ as the gradient estimator 
		\ENDFOR
		\STATE {\bfseries Output:} $\arg\max Q(\w^k)$ as the decision rule.
	\end{algorithmic}
\end{algorithm}

\begin{figure*}[t]
	\centering
	\begin{subfigure}{0.33\linewidth}
		\includegraphics[width=\linewidth]{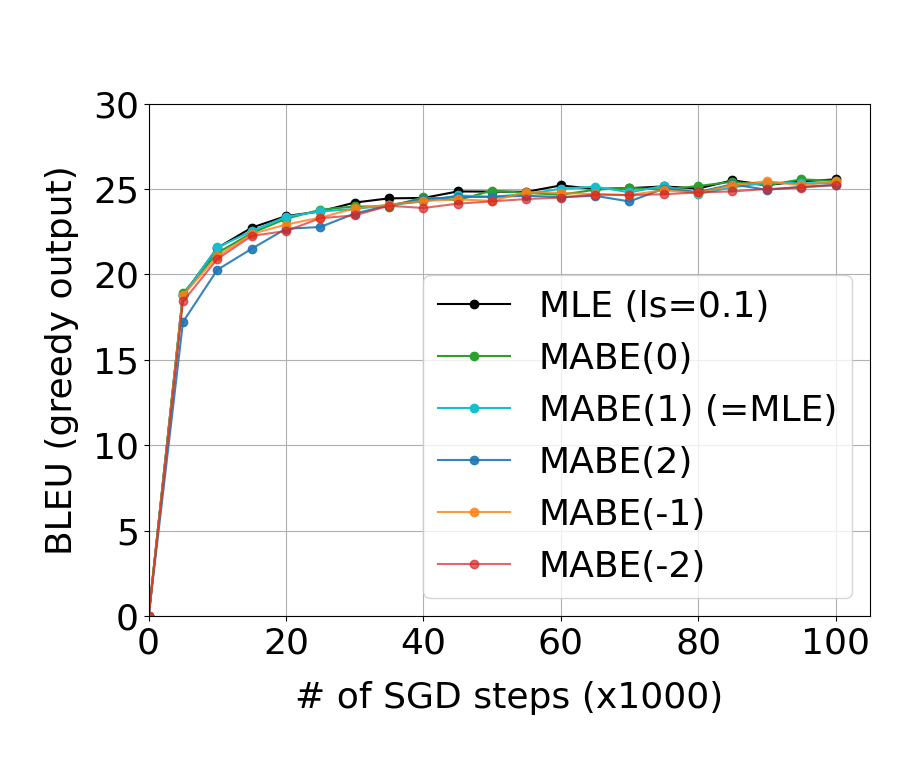}
		\caption{WMT'14 (English$\ra$German)}
	\end{subfigure}
	\hfill
	\begin{subfigure}{0.33\linewidth}
		\includegraphics[width=\linewidth]{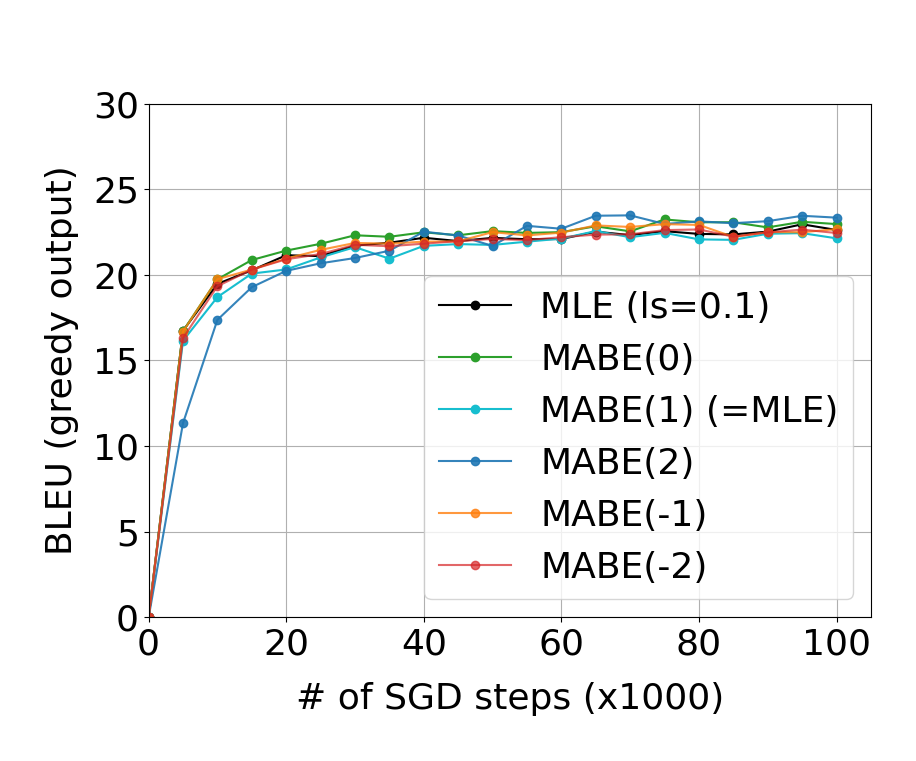}
		\caption{WMT'17 (Chinese$\ra$English)}
	\end{subfigure}
	\hfill
	\begin{subfigure}{0.32\linewidth}
		\includegraphics[width=\linewidth]{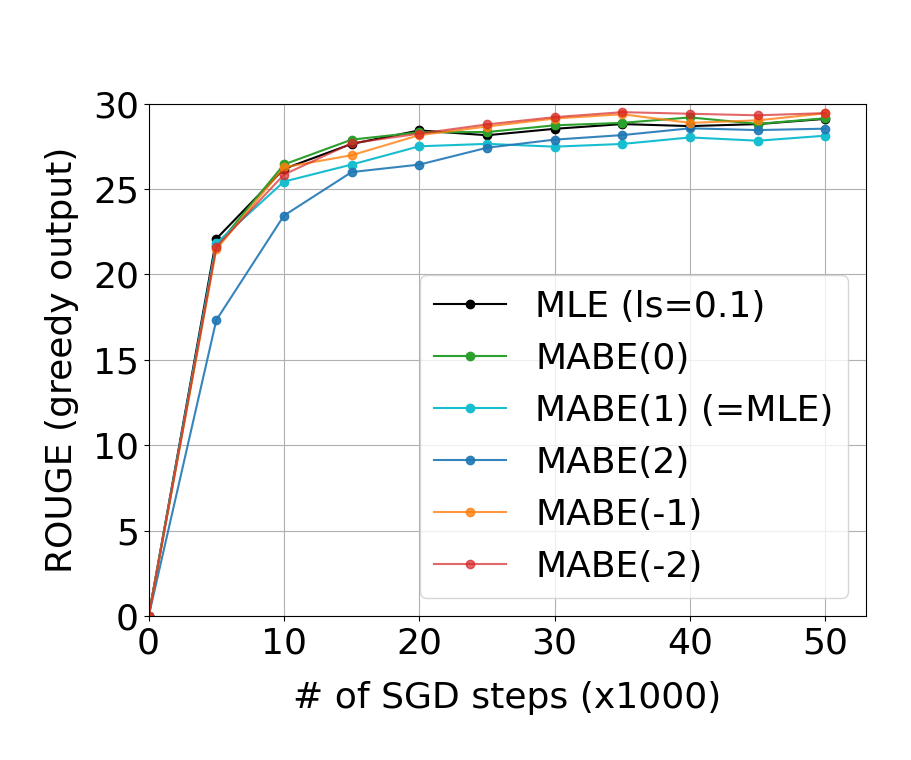}	
		\caption{CNN/DailyMail Summarization}
	\end{subfigure}
	\caption{SGD dynamic of $\Jmabe$ when the gradient is perturbed by the covariance term \eqref{cov}. Performance of the Q-greedy decision rule is evaluated on the test set every $5000$ SGD steps. BLEU and ROUGE are standard metrics for corresponding tasks.}
	\label{fig:GreedyQ}
\end{figure*}

Figure \ref{fig:GreedyQ} shows the learning curves of MABE($\lambda$) under five values of $\lambda$, ranging from $-2$ to $2$. Generally speaking, all the learning curves are similar, in all the three tasks being examined, not only in the end but almost throughout the training process. Performance of the perturbed variant MABE(1), a.k.a. MLE training, is slightly lower than the unperturbed variant MABE(0) (see Fig. \ref{fig:wmt14_mabe}, \ref{fig:wmt17_mabe}, \ref{fig:cnnmd_mabe} in the appendix for the numerical scores). The learning under $\lambda=2$ (which is 2x perturbed) was somewhat slower at the beginning, but managed to catch up with other variants in later stage of the learning. Importantly, MABE(1) -- a.k.a. the SGD-based MLE training of softmax-normalized neural networks -- does not seem to be anything uniquely different from the other ``non-probabilistic'' variants in the MABE($\lambda$) family. 
Under the greedy decision rule, its performance is generally similar to (often slightly lower than) that of the unperturbed variant MABE(0).


So far we have re-interpreted a classic statistical learning procedure (i.e. SGD-based MLE) as a kind of Q-learning algorithm (i.e. MABE($\lambda$)). Now we investigate the optimality of this Q-learning algorithm. Given that the performance of MABE($\lambda$) is similar under modest perturbation coefficient $\lambda$, in the following we focus on analyzing MABE(0), the learning dynamic without any perturbation. 
In the following we prove that when the Q-model is expressive enough, the global maximum of $\Jmabe$ is indeed an optimal Q-function (See the proof in Appendix \ref{sec:proof_mabe}).

\begin{theorem}
	\label{thm:mabe}
	Consider a tabular model $Q(a;\q)=\sum_{j=1}^d \indicator{a=j} \cdot q_j$, where the parameter vector $\vect{q}=(q_1 \dots q_d)$ directly encodes the utilities for each possible action $a\in \{1\dots d\}$. Let $\vect{p}=(p_1\dots p_d)$ be the softmax distribution induced by $\vect{q}$. 
	Let $\q^*$ be the Q-values that maximizes $J(\q) = \Exp[Y\sim \Ptrue]{Q(Y;\q)} - \Exp[A\sim \p]{Q(A;\q)}$, and $\p^*$ the softmax distribution by $\q^*$, then
	\eq[pqstar1]{
		\max_{a\in \supp(Y)}~ q^*_a ~~>~~ \max_{b\not\in \supp(Y)}~ q^*_b ~~+~~ 1
	}
	where $\supp(Y)$ is the support of $\Ptrue$ (which is thus the set of all desired actions), and moreover,
	\eq[pqstar2]{
		p^*_a \cdot (~ 1 + q^*_a - \E[]_{\p^*} [Q] ~) ~~=~~ \Ptrue[Y=a] \quad\quad,~~ \forall a 
	}
\end{theorem}
The function $J$ in Theorem \ref{thm:mabe} is an idealized form of the MABE objective $\Jmabe$, with the loss of parameterization in $Q(\w)$ being ignored. 
The inequality \eqref{pqstar1} in Theorem \ref{thm:mabe} suggests that $\q^*$, the global maximum of $J$, separates desired actions from undesired ones by at least a constant margin, so going greedy with $\q^*$ can provably avoid undesired actions. This fact established a strict optimality property for the global maximum of $\Jmabe$: Maximizing $\Jmabe$ over $\w$ guarantees to find the \emph{optimal} Q-function
\footnote{
	Recall that a Q-function is optimal if the greedy policy \eqref{greedyQ} finds optimal decisions (see Section \ref{sec:duality}).
} 
as long as the global maximum of $\Jmabe$ is covered by the parametric model $Q(\w)$.  

As MABE(0) is just a standard stochastic gradient process of $\Jmabe$ optimization, this optimality result (which is subject to model errors, data errors, and optimization errors) supports the soundness of the MABE family of Q-learning algorithms to the same strength as how the SGD-based MLE procedure has been justified in the probabilistic explanation of deep learning. 
In this way, we subsumed the SGD dynamic of MLE as a perturbed variant of a learning dynamic towards optimal Q-function (where the perturbation does not significantly affect the learning dynamic).

\begin{table*}[t]
	\small
	\centering
	\caption{Dual probabilities significantly improve pure sampling and MAP ($\pm$ gives $90\%$ confidence interval from $9$ trials).}
	\scalebox{0.95}{\begin{tabular}{l|rr|rr|rr}  
			\toprule
			& \multicolumn{2}{|c|}{WMT'14 en$\ra$de} &
			\multicolumn{2}{c|}{WMT'17 zh$\ra$en}	&
			\multicolumn{2}{c}{CNN / DailyMail} \\
			& 	sampling &	$|\text{beam}|=1024$ &	sampling &	$|\text{beam}|=1024$ &	sampling &	$|\text{beam}|=1024$ \\
			\midrule
			Softmax Probability & 5.4 $\pm$0.3	        & 8.6			& 8.3 $\pm$0.2			& 11.9		& 21.1 $\pm$0.1			& 20.1 \\
			Dual Probability 	& 20.0 $\pm$0.2			& 25.9 			& 17.7 $\pm$0.1			& 21.1 		& 27.9 $\pm$0.1	 		& 27.1  \\
			& (+14.6)				& (+ 17.3) 		& (+ 9.4) 		& (+ 9.2) 	& (+ 6.8) 		& (+ 7.1) \\
			\bottomrule
		\end{tabular}
	}
	\label{tab:stat-dataset}
\end{table*}

\section{The Dual Probabilities}
\label{sec:calibration}


In above we have been arguing that the empirical effectiveness of standard deep learning procedures can be better explained without interpreting the neural networks as probability models. In some cases, however, people may just want to have a probability model~\cite{2017:density}. Interestingly, our non-probabilistic theory entails a way to bring back the probabilistic semantic by transforming the learned Q-values (back) to probability estimations. 

Specifically, \eqref{pqstar2} in Theorem \ref{thm:mabe} gives a precise relationship between the Q-value of an action and the \emph{true} probability that the action is an desired one. The equation holds at the global maximum $\q^*$ of $\Jmabe$. In practice, the optimization is never exact for modern neural networks 
, yet we can still use the equation as a guidance to transform the Q-values obtained from MABE optimization to an approximation of the probabilities $\Ptrue$. 
Specifically, let vector $\q = (q_1\dots q_d)$ be the Q-values predicted by a Q-function for a given decision context, define
\eq[dual_prob]{
	\Pdual_i ~=~ \text{CLIP}\Big(~ p_i ~ (1 + q_i - \p \* \q) ~\Big) ~/~ Z
}
where $p_i = \text{softmax}_{[i]}(\q)$, $\text{CLIP}(x) \define \min(\max(0,x),1)$ trims the predictions to $[0,1]$, and $Z$ is the sum of the numerator in \eqref{dual_prob} across all $i$. The clipping and $Z$-normalization in \eqref{dual_prob} are not needed if $\q$ is exactly optimized.

We call the probability predictions by \eqref{dual_prob}, the \emph{dual probabilities} of the Q-values. Note that the dual probabilities $\Pdual$ are different from the predictions computed directly from the softmax transformation (which gives $\p$); the former ``calibrates'' the latter with a scaling factor $1+q_i+\p\* \q$. 

Empirically, we found that the dual probabilities \eqref{dual_prob} perform much better than the commonly used softmax probabilities, when both are used in probability-compatible decision rules, as Table \ref{tab:stat-dataset} shows (also see Appendix \ref{sec:wmt14en2de}, \ref{sec:wmt17zh2en} and \ref{sec:cnndm}). 

Taking WMT'14 en2de as example, translations by sampling the dual probabilities attain a BLEU score of $20.0$, which is a gain of $+14.6$ (or $+370\%$) over sampling with the traditional softmax probabilities (cf. Figure \ref{fig:mle_sampling_wmt14en2de}). The dual probability makes pure probability sampling a much more competitive decision rule now. 
Similarly, the dual probabilities also drastically improve the real-world performance of (approximate) probability maximization. For search with beam size $=1024$, for example, its BLEU score in WMT'14 en2de is lifted from $8.6$ to $25.9$, a gain of $+17.3$, and the score is higher than greedy's. 
Moreover, the dual probability of the empty output is now strictly zero on $2736$ of the $2737$ testing instances. In fact, the raw scaling factor $1+q_i-\p \* \q$ of the end-of-seq token was negative (thus was clipped to $0$) in all but one instances. 
It is only a pity that the model will also assign zero probability for most of the reference translations ($2614$ out of $2737$, which is less than the number for empty outputs though). On the other hand, the dual probability model is much more confident for self-generated outputs; see Figure \ref{fig:wmt14_dual} in Appendix \ref{sec:wmt14en2de}.

Similar gains in probability prediction and utilization can be observed in all tasks we examined. See Appendix \ref{sec:wmt14en2de}, \ref{sec:wmt17zh2en} and \ref{sec:cnndm} for more details. Overall, dual probability models exhibit much more reasonable behaviors than traditional softmax probability models.

Importantly, the dual probability formula \eqref{dual_prob} does not use any hyperparameter, and is derived from first principles. Recall that in Section \ref{sec:duality} we proposed to think of the Q-learning dynamic of \eqref{grad_lprob} as a dual process that simultaneously optimizes the Q-values \emph{and} the softmax probabilities. But now, in light of the advantage of the dual probabilities as observed in this section, it seems that the probability given by \eqref{dual_prob} is a more accurate probability model. As a result, if we say that the neural network is representing both utility and probability, the probability counterpart seems to be better represented by the probability given in \eqref{dual_prob}, instead of by the commonly recognized softmax probability.

\section{Conclusions}
\vspace{-0.05in}
\label{sec:discussion}

To summarize, we have demonstrated how the current practice of neural networks contradicts with its canonical probabilistic explanation in some complex decision tasks. This motivated us to develop an alternative explanation, in which the classic SGD-based MLE optimization process of softmax-normalized neural networks is interpreted as a supervised Q-learning algorithm (MABE(1)). 
Our utility-based theory is inherently free of the paradoxical probabilistic semantics, and yet can induce a dual probability space when needed, with significantly improved performance.

Based on results in this paper, one can either say that the neural network trained from ``SGD-based MLE optimization'' is modeling a utility function, whose theoretical optimality is characterized by Theorem \ref{thm:mabe}, or, one could also say that the neural network is indeed modeling a probability space, of not the softmax probability \eqref{softmax}, but of the dual probability \eqref{dual_prob}. 


Although this duality phenomenon may best manifest itself in sequential decision tasks,\footnote{In one-shot classification tasks, the MAP rule \eqref{map} degenerates to the greedy rule \eqref{greedy}, thus many problems observed in this paper will not show up, except that the inaccuracy of softmax probability estimations can still be observed~\cite{2017:calibration}.} we believe the conceptual implications of our duality theory may apply to all deep learning tasks because the probability interpretation of neural networks has been framed as a unified logical framework for all learning tasks. Our results challenged this mindset, and our theory provides a better unified framework to reason about deep neural networks.

\section*{Acknowledgments}
Authors of this paper would like to thank Rui Ding, Jingjing Xu, Shion Ishikawa, Liang Li, and Lihua Wang for giving many helpful comments on previous versions of this paper.

\bibliographystyle{plainnat}
\bibliography{rlnmt,rl}

\newpage
\appendix
\onecolumn

\section{On the Optimality of Probability Maximization}
\label{sec:map}
The Maximum-A-Posteriori principle is very intuitive in the sense that, given an input, if we have to make one and \emph{only one} output, then selecting the one that we think is most likely to be the expected output $Y$ is a quite natural idea. For a number of common performance metrics, this intuition is indeed formally supported by the provable optimality of maximizing the true probability $\Ptrue$.

Specifically, given a decision task with expected input $X$ and expected output $Y$, suppose the goal is to determine the actual output $A$ so as to maximize an \emph{expected utility} 
\eq[expected_utility]{
	\E_{X,A} [U(X,A)] = \sum_{\x} \P(\x) \sum_\a \P(\a|\x)~ U(\x,\a)
}
where
\eq[utility_avg]{
	U(\x,\a) = \sum_{\y} \Ptrue(\y|\x) ~\delta(\a,\y) 
}
and $\delta(\a,\y)$ is a similarity score between output $\a$ and output $\y$.

\textbf{Case 1}: 
Suppose $\delta$ is a binary score that simply measures if the actual output $A$ has correctly predicted $Y$ or not; that is, suppose $U(\a,\y)=\indicator{\a=\y}$. In this case the expected utility corresponds to the commonly used \emph{prediction accuracy}, and we have $U(\x,\a) = \sum_{\y} \Ptrue(\y|\x)~\indicator{\a=\y} = \Ptrue(\a|\x)$, so maximizing the probability $\Ptrue(\a|\x)$ is equivalent to maximizing the utility $U(\x,\a)$, given the input $\x$.

\textbf{Case 2}:  
Suppose for any given $\x$, the expected output $Y$ is deterministic, denoted by $\y_{\x}$. Equivalently, we are assuming that there exists a \emph{unique groundtruth} behind each observation; for example, given a picture $\x$, the object in that picture is always, say, a dog, no matter when or how many times the picture is observed. In this case, the distribution $\Ptrue$ degenerates and has $\Ptrue(\y_{\x}|\x)=1$, so we have $U(\x,\a)=\sum_{\y} \Ptrue(\y|\x)~\delta(\a,\y) = \delta(\a,\y_{\x})$. Now as long as $\delta$ is a similarity score, it is necessary that for any $\a$, $\delta(\a,\y_{\x}) \leq \delta(\y_{\x},\y_{\x})$ (i.e. $\y_{\x}$ is the one that is most similar to itself); in other words, the utility function $U(\x,\a)=\delta(\a,\y_{\x})$ attains its maximum at $\a=\y_{\x}$. On the other hand, as $\Ptrue(\a|\x)$ also attains its maximum at $\a=\y_{\x}$ (because $\Ptrue(\y_{\x}|\x)=1$), maximizing $\Ptrue(\a)$ is again equivalent to maximize $U(\x,\a)$ in this case (that is, when $Y$ is deterministic conditioned on $X$ and $\delta$ is a similarity score). 

\textbf{Case 3}:
In reality, the determinism assumption in Case 2 can be slightly relaxed, to the situation that the expected output $Y$ is not deterministic, but all the possible ``utterances'' of $Y$ has the same equivalent ``meaning'' (and that $\delta$ is a measure of \emph{semantic similarity}). For example, it is very common that given a question (e.g. ``what does the following English sentence mean in German?''), there is a definite answer at the semantic level, but this answer may admit multiple different yet equivalent expressions in natural language. In this case, there is still a unique groundtruth at the semantic level, and it is reasonable to say that an answer $A$ to the question should be judged affirmatively as long as it matches \emph{any} equivalent utterance of this unique groundtruth. 

It can be proved that for Case 3, delivering the ``most likely'' output $\Hmap$ is still optimal, for a similar reason as in Case 2. Specifically, given input $\x$, let $\supp(Y)$ be the support of the distribution $\Ptrue$ conditioned on $\x$. Suppose we replace the average-form utility \eqref{utility_avg} to a function that measures how similar the actual output $\a$ is with \emph{any} expected output $\y\in \supp(Y)$, that is, 
\eq[utility_max]{
	U(\x,\a) = \max_{\y\in \supp(Y)}~ \delta(\a,\y)  \quad\text{, where }~ Y\sim \Ptrue(\cdot|\x)
	.}
\eqref{utility_max} entails that $U(\x,\a)$ would attain its maximum at any $\a \in \supp(Y)$. This means the maximum point of $\Ptrue$, which must be within the support of $\Ptrue$ (i.e. within $\supp(Y)$), is necessarily a maximum point of the $U(\x,\a)$ in \eqref{utility_max} too. Note that the similarity measure $\delta$ in \eqref{utility_max} is general and needs not to be binary.

In summary, from above we see that maximizing the \emph{true} probability of $Y$ -- suppose we could do it -- is guaranteed to be exactly optimal either if the instance-wise utility is binary (Case 1), or if the groundtruth is unique (Case 2) or ``essentially unique'' (Case 3). As these conditions are quite common in practice, the optimality helps justify the widely-held conceptual reduction from optimal decision to probability estimation, and also underlie the widely-adopted decision rule of maximum a-posteriori which replaces the true probability $\Ptrue$ with the estimated a-posteriori probability $\P_{\w}$, with the hope that $\P_{\w}$, as a ``close'' approximation of $\Ptrue$, can still achieve ``good'' performance.

\section{On the Soundness of Probability Sampling}
\label{sec:sampling}

A probability model that specifies the distribution of the \emph{actual} output conditioned on given input is usually called a \emph{decision policy} in machine learning literature (particularly in reinforcement learning literature)~\cite{2018:RL}. As a special case, a deterministic policy maps each input to a definite output, and is also called a \emph{discriminant function} in statistics and statistical learning literature~\cite{2006:bishop}.
The probability sampling rule \eqref{sampling} asks the decision making agent to generate the actual output $A$ by sampling a learned distribution $\Pw$ of the expected output $Y$. In other words, $\Pw$ is used as a decision policy under the sampling rule.

When $\Pw=\Ptrue$, the actual output $A$ sampled from policy $\Pw$ would be identically distributed with the expected output $Y$ which follows $\Ptrue$. In this case if $Y$ is a ``target of prediction'' that is to be used in the utility function for similarity comparison, as in the case of \eqref{utility_avg} and \eqref{utility_max} in Section \ref{sec:map}, then an identically distributed $A$ to such $Y$ is not necessarily optimal. Instead, an optimal policy may generate the MAP output, which is deterministic, as discussed in Section \ref{sec:map}. 

However, the probability sampling rule enjoys a universal quality guarantee, regardless of the utility function, if the expected output $Y$ itself is the outcome of another policy. This can be easily observed from the expected utility formula \eqref{expected_utility}, where the performance of a policy depends only on the probability distributions induced by the policy. If $Y$ is considered an expected output, this is equivalent to say that $\Ptrue$ is an expected policy, in this case a policy giving the identical distribution $\Pw=\Ptrue$ is necessarily an expected policy too, by virtue of \eqref{expected_utility}. In particular, if $Y \sim \Ptrue$ is the outcome of an optimal policy, then the probability sampling rule \eqref{sampling} based on $\Pw=\Ptrue$ is necessarily an optimal decision rule too.

\begin{wrapfigure}{r}{0.3\linewidth}
	\centering
	\includegraphics[width=0.2\columnwidth]{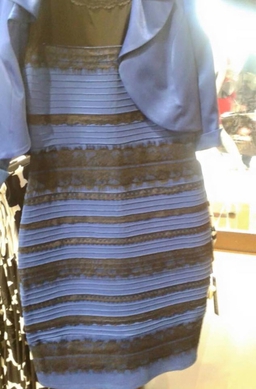}
	\caption{
	}
	\label{fig:dress}
\end{wrapfigure}

As a concrete example, consider the famous picture in Figure \ref{fig:dress}, and suppose the task is to predict the color of the dress in it by observing the picture (only). The ground truth is that the dress
 is in blue and black~\cite{2015:dress_news}, but average human beings are known to have divergent opinions on their observed color: $57\%$ saw the dress as blue and black, $30\%$ saw it as white and gold, $11\%$ saw it as blue and brown, and $2\%$ reported it as "other color", according to \citet{2015:dress_paper}. In this case, it seems that the optimal decision should match the unique and deterministic groundtruth (and following the MAP rule based on the human distribution indeed gives the optimal decision in this example). Yet, a decision policy that with probability $57\%$ outputs ``blue and black'' and with probability $30\%$ outputs ``white and gold'' etc. is \emph{as good as human beings} in predicting the groundtruth (for this particular picture), even though it may not be the optimal policy. 

Importantly, in many practices, the outputs in training data are indeed generated by randomly recruiting a group of human beings and letting them access to the same task inputs with what the AI/ML models would access to; the training data thus obtained does not represent the groundtruth, the true target of prediction, but is merely a sample of the ``human policy''. What matters here is only the overall conditional distribution (conditioned on the input) as represented by the training data, and reproducing the same distribution in AI/ML model's actual outputs should thus be considered as good (or as bad), no matter what the utility behind the human choices actually is. 

In summary, making the actual output $A$ identically distributed with $Y$ (or with as close distribution as possible) is a reasonable decision rule when the expected output $Y$ represents an imitation target at distributional level. In practice, this idea has been indeed widely adopted, and is known as \emph{behavior cloning} in imitation learning~\cite{2018:IL}. It is also the basic idea of \emph{language modeling} in NLP.



\section{Is the Training of Auto-regressive Models In Practice Really Following MLE Principle?}
\label{sec:mle}

The MLE principle is the theoretical foundation for the standard cross-entropy loss based training of auto-regressive models. This is explicitly documented in numerous literature. However, deep learning practice often deviate from its claimed theoretical rationale in some nuanced yet important ways, and some readers of this paper might think that the ``actual training algorithm'' for the auto-regressive models we studied is not optimizing the MLE objective \eqref{mle}, but is instead optimizing the token-level cross-entropy loss (for example, a reviewer of an older version of the paper is holding this position).

We point out that the objective function of "maximizing the (averaged) token probability", of "maximizing the (averaged) sequence probability", and of "maximizing the training data's probability" -- denoted by $J_{token}$, $J_{seq}$, and $J_{data}$ (resp.) -- these three objectives differ only in a constant factor. Specifically, for a training data consisting of $n$ $(\x[i],\y[i])$ pairs,
\eqm{
	J_{data} &\define \log \Pr[\w]{\{\y[1..n]\} | \{\x[1..n]\}} = \sum_{i=1}^n \sum_{t=1}^{T_i} \log \Pr[\w]{\yt[i]{t} | \x[i],\yp[i]{t}}  \\
	J_{seq} &\define \frac{1}{n} \sum_{i=1}^n \log \Pr[\w]{\y[i]|\x[i]} = \frac{1}{n} \sum_{i=1}^n \sum_{t=1}^{T_i} \log \Pr[\w]{\yt[i]{t} | \x[i],\yp[i]{t}} = J_{data} / n \\ 
	J_{token} &\define \frac{1}{\sum_{i=1}^n T_i} \sum_{i=1}^n \sum_{t=1}^{T_i} \log \Pr[\w]{\yt[i]{t} | \x[i],\yp[i]{t}} = J_{data} / \sum\nolimits_{i=1}^n T_i
}
where $n$ and all the $T_i$'s are constants. This means the three objectives must share the same \emph{optimization landscape} everywhere (including the same set of global optima, local optima, saddle points, etc.). Whenever one objective function is increased/decreased, the other two are necessarily increased/decreased too (and to the same rate).

Among the three, $J_{data}$ is exactly the MLE objective \eqref{mle}, which is in the sum-form. In order to estimate its gradient with a mini-batch, we have to turn it into average-form, e.g. to either $J_{seq}$ or $J_{token}$. In this, a notable implementation detail is that most popular training programs of auto-regressive models (huggingface, fairseq, tensor2tensor, etc.) sample the mini-batch in the unit of complete sequence, not in the unit of token. In other words, the mini-batch obtained in real-world training programs is an i.i.d. sample of \emph{only} the sequence-level loss $J_{seq}$. In particular, tokens in the same sequence are not independently sampled -- they are either included together or excluded together in the mini-batches. 

In summary, current training programs in practice are doing faithful SGD-based optimization for the MLE objective $J_{data}$, with the sequence-level loss $J_{seq}$ being its average-form surrogate for the sake of gradient estimation.


\section{Ordinal Utility, Cardinal Utility, and Value function in RL}
\label{sec:value}
The concept of utility was proposed and extensively studied in decision theory~\cite{1990:utility}, and is also central in microeconomics~\cite{2018:measure_utility}. In its cardinal interpretation, the utility of a decision is a real number that represents the ``value'' of the decision to a decision maker \emph{such that} the decision maker is said rational if it always chooses the option that maximizes the utility (or the expected utility when facing uncertainty). Such a cardinal utility function is assumed to (pre-)exist for a decision maker, and the decision maker can be classified as ``rational'' or ``irrational'' based on this utility function. 

As an example, the \emph{action-value} function in reinforcement learning~\cite{2018:RL}, usually denoted as a Q-function, can be seen as a cardinal utility function which is pre-defined by a reward function and the associated Markov Decision Process model, and a rational decision policy that maximizes its action-value under any state is an optimal policy. 

However, the cardinal interpretation of utility often suffers from measurability problem in practice~\cite{2018:measure_utility}. Given a specific decision maker, it is often hard to quantitatively measure its cardinal utility function (even if it exists). What is actually observable is the decisions eventually made by the decision maker. For example, this is exactly the case in the supervised learning setting as studied in this paper: the supervised training data only reveals the actual outputs chosen, not the utilities behind them. Moreover, real-world decision makers often has bounded rationality only, in which case the revealed decisions are not necessarily aligned with the cardinal utility function at all. 

For the above reasons, modern literature have largely shifted to estimating an \emph{ordinal utility} function from revealed preference of a decision maker. An ordinal utility function uniquely encodes a preference order over the choices by assigning each choice with a preferential score (a.k.a. the utility value). The absolute value of the score is not necessarily equal to any (hypothetical) cardinal utility, but it is solely the rank of the scores that is responsible for modeling the observed decision preferences revealed in data. This idea of ordinal utility was adopted in modern economics as a more practical approach to utility analysis(a.k.a. the so-called ``ordinal revolution''); see Chapter 5 of the book \cite{2018:measure_utility} for more information. 

We remark that even though the action-value functions in RL has a cardinal definition under the MDP framework, in practice the \emph{actual} Q-functions learned by RL algorithms often work as an ordinal utility function. For example, traditional Q-learning algorithms usually seek to iterate a parameterized Q-function toward the solution of a Bellman optimality equation. However, under nonlinear parameterization, the learned Q-function is known to diverge in general~\cite{2018:RL} and often predict Q-values that are far from the Bellman-optimal Q-function~\cite{2020:maxmin_qlearning,2015:ddpg}, yet it can still well support decision making in practice~\cite{2015:dqn}. Recently, new Q-learning algorithms have been proposed that do not seek to approximate the canonical Bellman-optimal Q-function at all~\cite{2021:iq} but focus on learning any Q-function with which the greedy policy is optimal~\cite{2022:lagrangian}. The optimal Q-function of this more general class is essentially an ordinal utility function that encodes the decision preference of an optimal policy without the cardinal constraint of being the Bellman-optimal Q-function.

In this paper we interpret the output of a modern neural network as an ordinal utility function as discussed above, and we explained why the ``apparent likelihood maximization'' procedure -- when applied to a big softmax-normalized model -- is approximately estimating a utility function and why this utility-based explanation better accounts for what we observed in experiments.

\section{Extended Discussion on Related Works}
\label{sec:existing_explanations}
The facts shown in Section \ref{sec:paradox}, that both maximizing and sampling the learned probability model perform poorly while going greedy or near-greedy with the local probabilities performed surprisingly well, and that the model assigns very low probabilities to desired outputs while giving much higher probabilities to undesired outputs, make one naturally wonder if the learned neural networks are really supporting decision making \emph{through} good probability modeling. In this section we discuss some existing explanations in the literature on this issue. 

%
%

\citet{2020:mbr} defended the probabilistic interpretation of the learned translation models by showing that the predicted probabilities $\Pw$ do match the groundtruth probabilities $\Ptrue$ in some statistics. They then attributed the pathological behavior of MAP output to the high entropy of the distribution being represented, arguing that when the most probable output has a probability as low as $1/10^4$, one should simply not trust the MAP rule. While this argument does make sense, we stress that the low predicted probability of the MAP output (or equivalently, the high entropy of $\Pw$) itself is suggesting that the learned distribution $\Pw$ is very different from the true distribution $\Ptrue$ as the latter \emph{should have been} highly concentrated. In fact, tasks like text translation or summarization fall in the category that the expected output is essentially unique, either directly at utterance level like in the specific WMT'14 dataset, or at the semantic level more generally; in this case the MAP output \emph{would} be optimal if the learned probability $\Pw$ \emph{were} indeed accurate about estimating $\Ptrue$ (see Section \ref{sec:map} for elaborations). In other words, the inadequacy of MAP output can only be \emph{caused} by the inaccuracy of probability modeling.

Many works do posit that the learned model deviates from the groundtruth distribution, and they seek to find out why. \citet{2016:ranzato} suggested that the distributional mismatch could be due to the inevitable discrepancy between the decision contexts that the model will encounter at training and testing time (in translation, for example, the decision context corresponds to the partial translations $\yp{t}$). \citet{2020:exposure} attributes the beam search pathology (observation (3) in Section \ref{sec:paradox_observations}) to this discrepancy. Subsequent studies, such as \cite{2019:bridging}, proposed methods to reduce this \emph{exposure bias} in order to facilitate better probability modeling. This line of works however did not explain why (near-)greedy decisions \emph{based on a model that is suffering from the exposure bias} can somehow lead to good empirical result. It appears that the model we obtained is not arbitrarily biased, but the bias \emph{happens to} best support a certain kind of decision rule (i.e. the greedy rule), across different tasks, different models, and different data sets.

\citet{2016:gnmt} observed that the beam search pathology 
is associated with the tendency to output shorter sentences under more extensive search. \citet{2019:stahlberg} confirmed that this \emph{length bias} is mostly due to errors in probability estimation, instead of to errors caused by the non-admissible search. On the other hand, \citet{2019:cohen} observed abnormal fluctuations of token-wise probabilities in pathological outputs, providing another factor associated to the search pathology. In addition, \citet{2019:sampling} found that the inadequacy of the sampled output (observation (1) in Section \ref{sec:paradox_observations}) is related to errors in the long tail of the distribution. All these works also complemented their discoveries with heuristic rules to prevent the search/sampling procedure from running into the identified pathological situations. These studies identified the failure patterns when the probability-based decision rules are getting wrong. But again, they did not shed too much light on why probability-\emph{incompatible} decision rules such as the greedy rule turn out to work much better. Note that the heuristics proposed in these works themselves have effectively made the resulted search or sampling procedure a deviation from the probability principles. These results leave it open for why we \emph{have to} deviate from well-established probability principles, either in the form of greedy decision or by some sort of heuristic-augmented search/sampling, to obtain competitive performance from the learned ``probability models''.



\citet{2020:uid} connected the pathological fluctuation of token-level probabilities as observed in \cite{2019:cohen} to a ``uniform information density (UID)'' hypothesis in cognitive science, offering a probability-based explanation to the effectiveness of greedy output. However, the precise mathematical expression of the UID hypothesis itself is subject to different interpretations~\cite{2021:uid}. For example, \citet{2020:uid} examined a number of different regularization terms, all considered as a form of UID regularization; it turns out that the two ``purest UID regularizers'' performed the worst, while a ``greedy UID regularizer'' (Eq.11 of the paper, which literally penalizes for deviating from the greedy output) performs the best. It is then subject to debate regarding if the greedy UID regularizer here has facilitated preference to UID outputs or directly to greedy outputs. In contrast, in this paper we seek to develop a mathematically clear and non-probabilistic account to explain the effectiveness of greedy outputs.


\section{Proofs}
\label{sec:proofs}

\subsection{Proof of Proposition \ref{thm:grad_greedyq}}
\label{sec:proof_mle}

\begin{recap}[Proposition \ref{thm:grad_greedyq}]
	Given input $\x$, output $\y=\seq{\yt{1} \dots \yt{T}}$, and parametric model $Q(\w)$, 
	we have
	\eqm{
		\pde{\log \Pr[\w]{\y|\x}}{w_j}  =  
		\pde{\Jmabe}{w_j} (\w,\x,\y) ~+~ \sum_{t=1}^T \qcov{t}{j}  \quad\quad,~~ \forall j
	}
	where~ $\qcov{t}{j} \define \Cov[\Hpi_t \sim P_t]{Q_t(\Hpi_t) ~,~ \pde{Q_t}{w_j}(\Hpi_t)}$
	\eqa{
		=~&	\sum_a P_t(a) \cdot \Big( Q_t(a) - \sum_{b} P_t(b) Q_t(b) \Big) 
		\cdot \Big( \pde{Q_t}{w_j}(a) - \sum_{b} P_t(b) \pde{Q_t}{w_j}(b) \Big) \hspace{0.55in}
	}
	\vspace{-0.1in}
	\squishlist
	\item $\P_{\w}$ is the softmax distribution induced by $Q(\w)$ as prescribed by \eqref{prob_y} and \eqref{softmax},
	\item $P_t(a)$ denotes $\Pr[\w]{a|\x,\yp{t}}$, the probability that the token at step $t$ is $a$, according to $\Pw$,
	\item $Q_t(a)$ denotes $Q(a|\x,\yp{t};\w)$, the utility of outputting token $a$ at step $t$, according to $Q(\w)$,
	\item $\pde{Q_t}{w_j}(a)$ denote $\pde{Q}{w_j}(a|\x,\yp{t};\w)$, the partial derivative of $Q(a|\x,\yp{t};\w)$ at step $t$,
	\squishend
\end{recap}

\begin{proof}
	From \eqref{grad_lprob} (and \eqref{prob_y}), we have that for any component $j$ of the model parameter $\w$,
	\eq{
		\pde{\log \Pr[\w]{\y|\x}}{w_j} = \sum_{t=1}^T \Big(~ 
		\pde{Q_{[\yt{t}]}}{w_j}(\x,\yp{t};\w) - 
		\sum_a ~ f_{[a]}(\x,\yp{t};\w) ~ \pde{Q_{[a]}}{w_j}(\x,\yp{t};\w) 
		~\Big)
		.}
	By definition of $\Jmabe$ (i.e. \eqref{J_mabe}),
	\eqm{
		\pde{\Jmabe(\w,\x,\y)}{w_j} 
		=~&	\pdbyd{w_j}	\sum_{t=1}^{T}~ \Big(~ 
		Q_{[\yt{t}]}(\x,\yp{t};\w) - \sum_{a} f_{[a]}(a|\x,\yp{t};\w) Q_{[a]}(\x,\yp{t};\w) 
		~\Big) \\
		=~&	\sum_{t=1}^{T}~ \Bigg(~ 
		\pde{Q_{[\yt{t}]}}{w_j}(\x,\yp{t};\w) - 
		\sum_a f_{[a]}(\x,\yp{t};\w) ~ \pde{Q_{[a]}}{w_j}(\x,\yp{t};\w) \\
		&	\quad\quad\quad - \sum_a Q_{[a]}(\x,\yp{t};\w) ~ \pde{f_{[a]}}{w_j}(\x,\yp{t};\w) ~\Bigg) \\
		=~&	\pde{\log \Pr[\w]{\y|\x}}{w_j} - \sum_t \sum_a Q_t(a) ~ \pde{P_t}{w_j}(a)
	}
	So now we only need to prove that 
	\eq[covt_proof1]{
		\sum_a Q_t(a) ~ \pde{P_t}{w_j}(a) = \Cov[\Hpi_t \sim P_t]{Q_t(\Hpi_t) ~,~ \pde{Q_t}{w_j}(\Hpi_t)}
		.}
	At the right-hand side of \eqref{covt_proof1}, $Q_t(\Hpi_t)$ and $\pde{Q_t}{w_j}(\Hpi_t)$ are two random variables defined on top of the sample space of $\Hpi_t \sim P_t$. Recall that for any random variables $X$ and $Y$, $\cov [X,Y] \define \E[] [(X-\E[][X])(Y-\E[][Y])] = \E[] [X(Y-\E[][Y])]$, thus
	\begin{align}
		\Cov[\Hpi_t \sim P_t]{Q_t(\Hpi_t) ~,~ \pde{Q_t}{w_j}(\Hpi_t)} 
		=~&	\Exp[\Hpi_t \sim P_t]{
			Q_t(\Hpi_t) ~ \Big(~ 
			\pde{Q_t}{w_j}(\Hpi_t) - \E[\Hpi_t\sim P_t] \big[~ \pde{Q_t}{w_j}(\Hpi_t) ~\big] 
			~\Big)
		} \nonumber\\
		=~&	\sum_a P_t(a) ~ Q_t(a) ~ \Big(~
		\pde{Q_t}{w_j}(a) - \sum_b P_t(b) \pde{Q_t}{w_j}(b)
		~\Big) \label{covt_proof2}\\
		=~&	\sum_a P_t(a) ~ Q_t(a) ~ \pde{\log P_t(a)}{w_j} \label{covt_proof3}\\
		=~&	\sum_a Q_t(a) ~ \pde{P_t(a)}{w_j} \nonumber
	\end{align}
	Note that from \eqref{covt_proof2} to \eqref{covt_proof3} we have utilized the equation \eqref{grad_lprob} again.
\end{proof}

\subsection{Proof of Theorem \ref{thm:mabe}}
\label{sec:proof_mabe}
\begin{recap}[Theorem \ref{thm:mabe}]
	Consider a tabular model $Q(a;\q)=\sum_{j=1}^d \indicator{a=j} \cdot q_j$, where the parameter vector $\vect{q}=(q_1 \dots q_d)$ directly encodes the estimated utilities for each possible action $a\in \{1\dots d\}$. Let $\vect{p}=(p_1\dots p_d)$ be the softmax distribution induced by $\vect{q}$. 
	Let $\q^*$ be the Q-values that maximizes $J(\q) = \Exp[Y\sim \Ptrue]{Q(Y;\q)} - \Exp[A\sim \p]{Q(A;\q)}$, and $\p^*$ the corresponding softmax distribution, 
	\renewcommand\labelenumi{(\theenumi)}
	\begin{enumerate}
		\item for any action $a \in \{1\dots d\}$, 
		\eq{
			p^*_a \cdot (~ 1 + q^*_a - \E[]_{\p^*} [Q] ~) ~~=~~ \Ptrue[Y=a]
		}
		\item let $\supp(Y)$ be the support of $\Ptrue$ (which is thus the set of all desired actions),
		\eq{
			\max_{a\in \supp(Y)}~ q^*_a ~~>~~ \max_{b\not\in \supp(Y)}~ q^*_b ~~+~~ 1
		}
	\end{enumerate}
\end{recap}

\begin{proof}
	We first prove Conclusion (1). 
	Applying Proposition \ref{thm:grad_greedyq} to $J(\q)$ -- which corresponds to a special case of $\Jmabe$ with $T=1$ (single step) and $|\supp(X)|=1$ (single input) -- yields
	%
	\eq[mabe_proof1]{
		\pde{J}{q_j} = 
		\Exp[Y\sim \Ptrue]{\pde{Q}{q_j}(Y;\q)} - \Exp[A\sim \p]{\pde{Q}{q_j}(A;\q)} - \Cov[A\sim \p]{ Q(A;\q),\pde{Q}{q_j}(A;\q) } 
	}
	Since $\pde{Q}{q_j}(a;\q) = \indicator{a=j}$, we have that for \emph{any} random variable $Z$,
	\eq[mabe_proof2]{
		\E[]_{Z} \Big[ \pde{Q}{q_j}(Z;\q) \Big] ~=~ \sum_z \Pr{Z=z} ~\pde{Q}{q_j}(z;\q) ~=~ \sum_z \Pr{Z=z}~ \indicator{a=j} ~=~ \Pr{Z=j}
		.}
	Applying \eqref{mabe_proof2} to \eqref{mabe_proof1}, yields
	\eqm{
		\pde{J}{q_j} 
		&= 	\Ptrue(j) - p_j - \Cov[A\sim \p]{ Q(A;\q),\pde{Q}{q_j}(A;\q) } \\
		&=	\Ptrue(j) - p_j + \E_{\p} [Q] \cdot \E_{\p} [\pde{Q}{q_j}] - 
		\E_{\p} [Q\cdot \pde{Q}{q_j}] \\
		&=	\Ptrue(j) - p_j + \E[] [Q] \* p_j - \E_{\p} [Q\cdot \pde{Q}{q_j}]
	}
	With the same the argument as in \eqref{mabe_proof2}, we similarly have $\E_{\p} [Q\cdot \pde{Q}{q_j}] = \sum_a p_a \* q_a \* \indicator{a=j} ~=~ p_a \* q_a$, thus
	\eq[pqstar3]{
		\pde{J}{q_j} = \Ptrue(j) - p_j + \E[]_{\p} [Q] \* p_j - p_j \* q_j
	}
	At $\p^*$ and $\q^*$, $\pde{J}{q_j} = 0$ for all $j$, so
	\eq[pqstar]{
		p^*_j \cdot (~ 1 + q^*_j - \E[]_{\p^*} [Q] ~) ~=~ \Ptrue(j) \quad,~~ \forall j \in \{1\dots d\}
	}
	which gives Conclusion (1). 
	
	To derive Conclusion (2), we will prove a stronger result, that 
	\begin{proposition} 
		\label{prop:mabe}
		A $\q^*$ that maximizes $J$ will assign the same utility, $\E[\p^*] [Q]-1$, to every undesired action $b\not\in \supp(Y)$.
	\end{proposition} 
	What immediately follows Proposition \ref{prop:mabe} is that there must be at least one desired action $a\in \supp(Y)$ such that $q^*_a > \E[]_{\p^*}[Q]$ (as otherwise no action would have utility above the \emph{averaged} utility $\E[]_{\p^*}[Q]$), and therefore, $\max_{a\in \supp(Y)}~ q^*_a ~~>~~ \E[]_{\p^*}[Q] ~~=~~ \max_{b\not\in \supp(Y)}~ q^*_b + 1$.
	
	Now we prove Proposition \ref{prop:mabe}.
	Consider an arbitrary undesired action $b\not\in \supp(Y)$. For any such $b$, $\Ptrue(b)=0$. As a result, the left-hand side of \eqref{pqstar} must be zero too, for $j=b$; that is, $p^*_{b} \cdot (1+q^*_{b} - \E[]_{\p^*}[Q]) = 0$. So there can be only two possibilities: either $1+q^*_b-\E[]_{\p^*}[Q] = 0$ which is exactly what Proposition \ref{prop:mabe} is asserting, or $p^*_b = 0$. In the following we show that $p^*_b = 0$ is impossible, even as a limit.
	
	Strictly speaking, $p^*_b$ cannot be exactly zero simply because $\p^*$ is a softmax distribution with finite logits. But one may wonder the possibility that a $\p^*$ with $p^*_b=0$ is a \emph{supremum} of $J$ in the limit; in that case the optimization of $J(\q)$ (i.e. the MABE optimization) may update $q^*_b$ towards $-\infty$ (although never reaches it), thus breaks Proposition \ref{prop:mabe}.
	\footnote{
		Note that in this case $\E[]_{\p^*}[Q]=\sum_i p^*_i q^*_i$ is still finite, thus well defined, because $\lim\limits_{q^*_b \ra -\infty} p^*_{b} q^*_{b} = 0$.
	}
	
	It turns out that such a supremum with $q^*_b=-\infty$ cannot exist for the $J$ as defined. Specifically, consider the process of taking an arbitrary $\q$ with $q_b = \E[][Q] - 1$ then decreasing $q_b$ down toward $-\infty$ (while keeping all other $q_{j\neq b}$ fixed). As $q_b < \E[][Q]-1$ throughout this process, we have $1 + q_b - \E[][Q] < 0$ all the time, and thus by \eqref{pqstar3}, $\pde{J}{q_b} = \Ptrue(b) - p_b (1+q_b-\E[][Q]) > 0$ throughout the process. Since $q_b$ is decreasing, $J$ must be also decreasing due to the positive derivative. Therefore, $J$ cannot attain a maximum (or supremum) at $q_b = -\infty$, not even a local maximum/supremum. 
\end{proof}

\begin{figure}[H]
	\begin{minipage}{1.0\linewidth}
		\centering
		\includegraphics[width=0.4\linewidth]{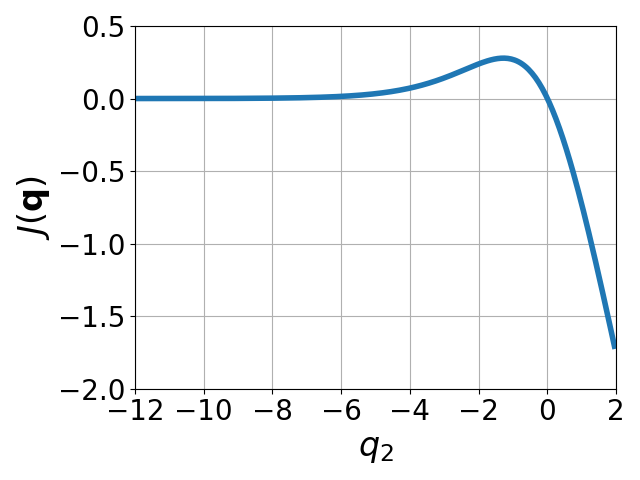}
		\caption{Shape of the MABE objective $J(\q)$, for $\q=(q_1,q_2)$. 
		}
		\label{fig:degenerate_q}
	\end{minipage}
\end{figure}

As an example, Figure \ref{fig:degenerate_q} illustrates the function $J$ when there are only two actions (i.e. $d=2$) and assume action $1$ is the expected/desired one. In this case $\q=(q_1,q_2)$, and Figure \eqref{fig:degenerate_q} shows the cross section at $q_1=0$ for the 2D function $J(\q)$ (the shape is the same at all $q_1$'s). The global maximum of $J(\q)$ is attained at $q_2 = \E[\p] [\q] - 1$. Notably, we see that the MABE objective function $J$ is non-convex, and yet it has a unique \emph{local} maximum point. 

\section{More Experimental Results}
\label{sec:experiments}

In this section we present our experimental results in details, task by task.

\subsection{Translation-WMT14en2de}
\label{sec:wmt14en2de}

The WMT'14 NewsTest English$\ra$German (en2de) task asks to translate English sentences in news articles into German sentences, and is among the most-used public benchmarks in machine translation. In this task, the expected input $X$ is an English sentence following distribution defined by the training data, the expected output $Y$ is a sequence of German tokens, ending with a special End-Of-Sentence(EOS) token. EOS is generated either as part of the actual output (by the learned model), or is forced by the system when the translation is $2$x long compared with the source sentence.

The training dataset~\footnote{https://nlp.stanford.edu/projects/nmt/} consists of $4.5$ million sentences of translation examples, and the testing dataset consists of $2737$ examples. The data was pre-processed and post-processed using the BPE tokenizer~\cite{2016:bpe} provided by YouTokenToMe~\footnote{https://github.com/VKCOM/YouTokenToMe}, with shared vocabulary of size $37000$. We used SacreBLEU~\cite{2018:sacrebleu} to calculate the BLEU scores. 

We first trained the standard TransformerBase neural network~\cite{2017:transformer} for $100,000$ SGD steps with the standard cross-entropy loss, which corresponds exactly to the MLE procedure \eqref{mle} as discussed in the paper. We followed the same hyperparameter setting recommended in \cite{2017:transformer}, which is known to achieve a BLEU score around $27.3$ under a near-greedy decision rule. 
A dropout rate of $0.1$ and labeling smoothing of $0.1$ are applied, again as recommended by \cite{2017:transformer}.

The learned neural network was then used to power a number of different decision rules: 
\begin{itemize}
	\item Temperature-regulated Sampling: A generalization of the pure sampling rule \eqref{sampling}, in which the sampling probabilities are scaled by a temperature parameter $\beta$, with $\P[A_t=\yt{}|\x,\yp{t}] \propto (\Pw[Y_t=\yt{}|\x,\yp{t}])^{1/\beta}$. When $\beta=1$, the temperature-regulated sampling implements exactly the probability sampling rule \eqref{sampling}. With $\beta$ tuned toward $0$, the parameterized sampling deviates from the genuine probability predictions, and is biased more and more toward near-greedy outputs. The greedy rule corresponds to $\beta=0$. 
	\item Greedy Decoding: The greedy decision rule \eqref{greedy}.
	\item Beam Search: The standard and \emph{vanilla} version of beam-search decoding, which uses the estimated probability as the heuristic score of a partial output in the search (see Algorithm 1 in \cite{2019:stahlberg}).
\end{itemize}

The translation performance on the testing set is reported in Figure \ref{fig:wmt14_mle}. The figure also shows the estimated probabilities according to the softmax probability model (the sentence-level probability of the translation for each testing instance was estimated, then their logarithms with base $10$ were averaged over all testing instances). As discussed in Section \ref{sec:paradox_observations}, these results exhibit counter-intuitive behaviors if we think of the learned neural network as a probability model.

In comparison, the translation performance and the estimated probabilities make much more sense if we instead use the dual probabilities as prescribed by \eqref{dual_prob}, as Figure \ref{fig:wmt14_dual} shows: The sampling output based on dual probabilities is still not the best (as expected, see Section \ref{sec:sampling}), but is much more reasonable now. Going greedy with the dual probabilities gives better outputs than sampling (25.6 vs 20.0), and in fact is giving the same output with the greedy ones under softmax probabilities (this is expected as both probability models are order preserving). The beam search result, approximately representing the MAP output, is further slightly better the greedy ones (25.9 vs 25.7) under the dual probabilities, which again aligns with the expectation that truly maximizing the probabilities should help with the performance (instead of hurting it, as in the case of softmax probability). 

In terms of probability estimation, the dual probability model is generally less perplexed than the softmax model, as Figure \ref{fig:wmt14_dual}(right) shows. The likelihood of empty output (which is zero) is also correctly estimated by the dual probabilities now. It is only a pity that the model will also assign zero probability for most of the reference translations ($2614$ out of $2737$, which is less than that for empty outputs though). In fact, the dual probability model will assign reasonable likelihood to most \emph{tokens} in the reference translations, but may only occasionally judge some tokens as ``impossible''; however, once there is a single token is judged so in a reference translation, the probability of the whole sentence becomes zero. We tend to think of this as a fragile nature of the probability-based method in general.  

Finally, Figure \ref{fig:wmt14_mabe} shows the learning curves of different MABE($\lambda$) variants in WMT'14 en2de, which complements Figure \ref{fig:GreedyQ}(a) of Section \ref{sec:q-learning} with more numerical details. All variants use the same hyperparameter setting with the aforementioned MLE training, except that label smoothing is disabled (as it is incompatible to value learning). Each point in the figure gives the test-set BLEU score of model trained by a given algorithm variant for a given number of SGD steps. The results are averaged over four random seeds.

We see that all algorithms demonstrate similar learning curves, with MABE(2) slighly left behind at the beginning stage. MABE(0), the unperturbed variant, appears to perform slightly better than other perturbed variants.

\subsection{Translation-WMT17zh2en}
\label{sec:wmt17zh2en}

The WMT'17 Chinese$\ra$English (zh2en) task asks to translate news articles in Chinese into English. The expected input $X$ is a Chinese sentence (distribution defined by the training data), and the expected output $Y$ is a sequence of English tokens, again ending with the EOS token which is either generated or forced when the output reaches $2$x long than the input. As the two languages are more distinct, it is generally considered a harder translation task than WMT'14 en2de. 

The raw WMT'17 zh2en training data contains 25 million sentences of translation examples from three sources: News Commentary, UN Parallel Corpus and CWMT Corpus.
\footnote{https://www.statmt.org/wmt17/translation-task.html\#download} 
We cleaned the raw data following the steps described in \cite{2018:achieving} (with slight difference in parameter details):
\begin{itemize}
	\item Sentences with illegal characters (such as URLs, characters of other languages) and empty sentences are removed.
	\item Duplicate translation examples are dropped.
	\item Both the source and target sentences should contain at least 3 words and at most 80 words.
	\item Chinese sentences without any Chinese characters are discarded.
\end{itemize}
The final training data set consists of about 21 million sentence-pairs. The testing set \emph{newstest2017} was left intact. For Chinese data, we adopting the Jieba tokenizer\footnote{https://github.com/fxsjy/jieba} before the byte pair encoding (BPE). English sentences are tokenized using the scripts provided in Moses before using BPE. The BPE vocabularies for Chinese and English are generated separately, each with a merge-operation budget of 32000. The generated Chinese and English vocabulary contains 50K and 33K sub-word tokens, respectively.

We conducted the same set of experiments in WMT'17 zh2en as we did in WMT'14 en2de: We trained a TransformerBase neural network (8 heads for each multi-head module in both encoder and decoder layers; 512 for the dimensions of input and output layers, and 2048 for the inner feed-forward layers), first with the standard MLE loss then with the MABE($\lambda$) losses, for $100,000$ SGD steps each. The hyperparameter setting is identical across all losses (optimizer and learning rate settings followed \cite{2017:transformer}, training batch size is roughly 36,000 English tokens, drop-out rate is 0.3), except that the label smoothing weight is 0.1 for MLE (0.0 for MABE variants, including MABE(1)). BLEU calculation is by SacreBLEU.

Figure \ref{fig:wmt17_mle} illustrates the behaviors of the MLE model under various decision rules. We see that Probability sampling and maximization (approximated by beam search with beam size $=1024$) failed in WMT'17 zh2en too, achieving only $8.3$ and $11.9$, respectively. In comparison, greedy outputs reaches $22.6$, which is only $0.8$ BLEU lower than the best solution (beam search with beam size $=4$). All the four observations discussed in Section \ref{sec:paradox_observations} occurred in WMT'17 zh2en too. Similarly, the dual probabilities dramatically improve the performance of sampling and MAP in WMT'17 zh2en, to the levels that match the expectations much better, as Figure \ref{fig:wmt17_dual} shows.

Trends in the learning curves of different MABE variants (Figure \ref{fig:wmt17_mabe}) also very much resembled what we saw in WMT'14 en2de. Each data point is the averaged result over four trials.
There is no clear ``winner'' or ``loser''-- MABE(2), which receives doubled perturbation, is again slower in the learning progress at the beginning, but it quickly catches up and slightly exceeds other variants around the end of the training. The experimental results once again support our main hypothesis that the covariance-based perturbation in MABE($\lambda$) does not make significant difference under modest $\lambda$.

\subsection{Summarization-CNN/DM}
\label{sec:cnndm}

The CNN/DailyMail Summarization task asks to generate abstracts for news articles collected from CNN and DailyMail. The excepted input $X$ is a full news article in English, and the expected output $Y$ is a (much shorter) sequence of English tokens, ending with EOS. EOS is forced if the length of the output reaches $1024$. Comparing with translation tasks, the CNN/DailyMail task has about $30$x larger inputs and about $3$x larger outputs.

The datasets were prepared by following the procedure used by \citet{2020:bart}. We first downloaded the raw CNN/DailyMail dataset~\footnote{https://cs.nyu.edu/$\sim$kcho/DMQA/}, then pre-processed the data (including the train-dev-test split) using scripts~\footnote{https://github.com/abisee/cnn\-dailymail} from \cite{2017:cnndm}. 
Then following the code repository~\footnote{https://github.com/pytorch/fairseq/blob/main/examples/bart/README.summarization.md} of \cite{2020:bart}, we used the GPT-2's BPE model to convert the pre-processed data into tokenized sequences over a BPE vocabulary. The final training data consists of nearly $0.3$ million article-abstract pairs, and the testing set consists of $2000$. ROUGE, the standard document-summarization metric, was employed for performance measurement. More specifically, the ROUGE score reported below is the arithmetic mean of the F1 scores of ROUGE-1, ROUGE-2, and ROUGE-L. 

The TransformerBase neural network in this task has 8 attention heads, 768 dimensions for input and output layers, and 2048 dimensions for the inner feed-forward layers. We trained the neural network on an A100 GPU, again using the cross-entropy (=MLE) loss and the MABE($\lambda$) loss with $\lambda=-2,-1,0,1,2$. Each training lasts for $50,000$ gradient steps, and each step is based on a mini-batch of about 64 instances. Drop-out rate is 0.1, and label smoothing is 0.1 for MLE; no label smoothing for MABE variants. All the decision rules used at test time are identical to the ones used in translation experiments.

From Figure \ref{fig:cnnmd_mle} we see that the probability sampling and maximization rules perform better in CNN/DailyMail summarization, but still have wide gap in performance compared with the greedy rule ($8.0$ and $9.0$ ROUGE scores lower, respectively). Once again, the dual probability model largely filled the gap here, as Figure \ref{fig:cnndm_dual} shows. 

Figure \ref{fig:cnnmd_mabe} shows the learning curves of different MABE variants on CNN/DailyMail. Each data point is the mean value over two trials.
It appears that positive perturbations (MABE(1) and MABE(2)) lead to slightly lower result, while negative perturbations (MABE(-1) and MABE(-2)) lead to slightly higher result. The gap between the unperturbed variant MABE(0) and and the MLE-equivalent variant MABE(1) is relatively larger in this task (around $1$ ROUGE score). 

\newpage

\begin{figure}[t]
	\begin{minipage}{0.5\linewidth}
		\includegraphics[width=\linewidth]{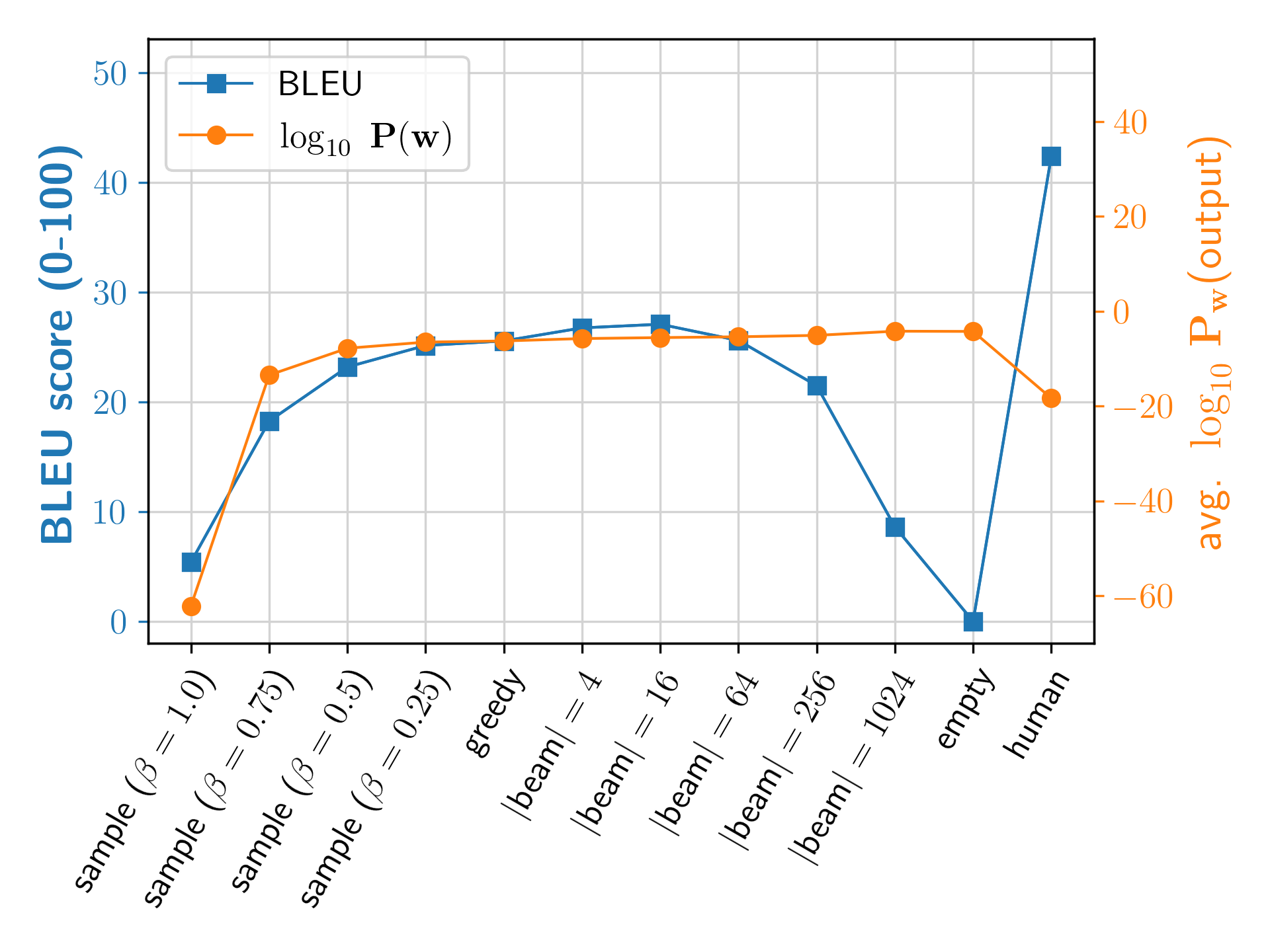}
\end{minipage}
\hfill
\begin{minipage}{0.4\linewidth}
	\centering
	\scalebox{0.95}{\begin{tabular}{|l|r|r|}
			\hline
			& BLEU 	& $\log_{10} \Pw$ \\
			\hline
			sampling 			& 5.4 & -62.2 \\
			\hline
			$\beta=0.75$ 		& 18.3 & -13.5 \\
			$\beta=0.5$ 		& 23.2 & -7.8 \\
			$\beta=0.25$ 		& 25.1 & -6.5 \\
			\hline
			greedy 				& 25.6 & -6.3 \\
			\hline
			$|\text{beam}|=4$ 	& 26.8 & -5.8 \\
			$|\text{beam}|=16$ 	& 27.1	& -5.6 \\
			$|\text{beam}|=64$ 	& 25.6 & -5.4 \\
			$|\text{beam}|=256$	& 21.5 & -5.1 \\
			$|\text{beam}|=512$	& 16.2 & -4.7  \\
			$|\text{beam}|=1024$& 8.6 & -4.2 \\
			\hline
			empty 				& 0.0 	& -4.3 \\
			\hline
			expected 			& 42.4~\footnote{See calculation method in Section \ref{sec:ref_perf}.} & -18.4 \\
			\hline
	\end{tabular}}
\end{minipage}
\caption{On WMT'14 en$\ra$de, MLE model exhibits paradoxical behaviors.}
\label{fig:wmt14_mle}
\end{figure}

\begin{figure}[H]
\begin{minipage}{0.4\linewidth}
	\centering
	\includegraphics[width=\linewidth]{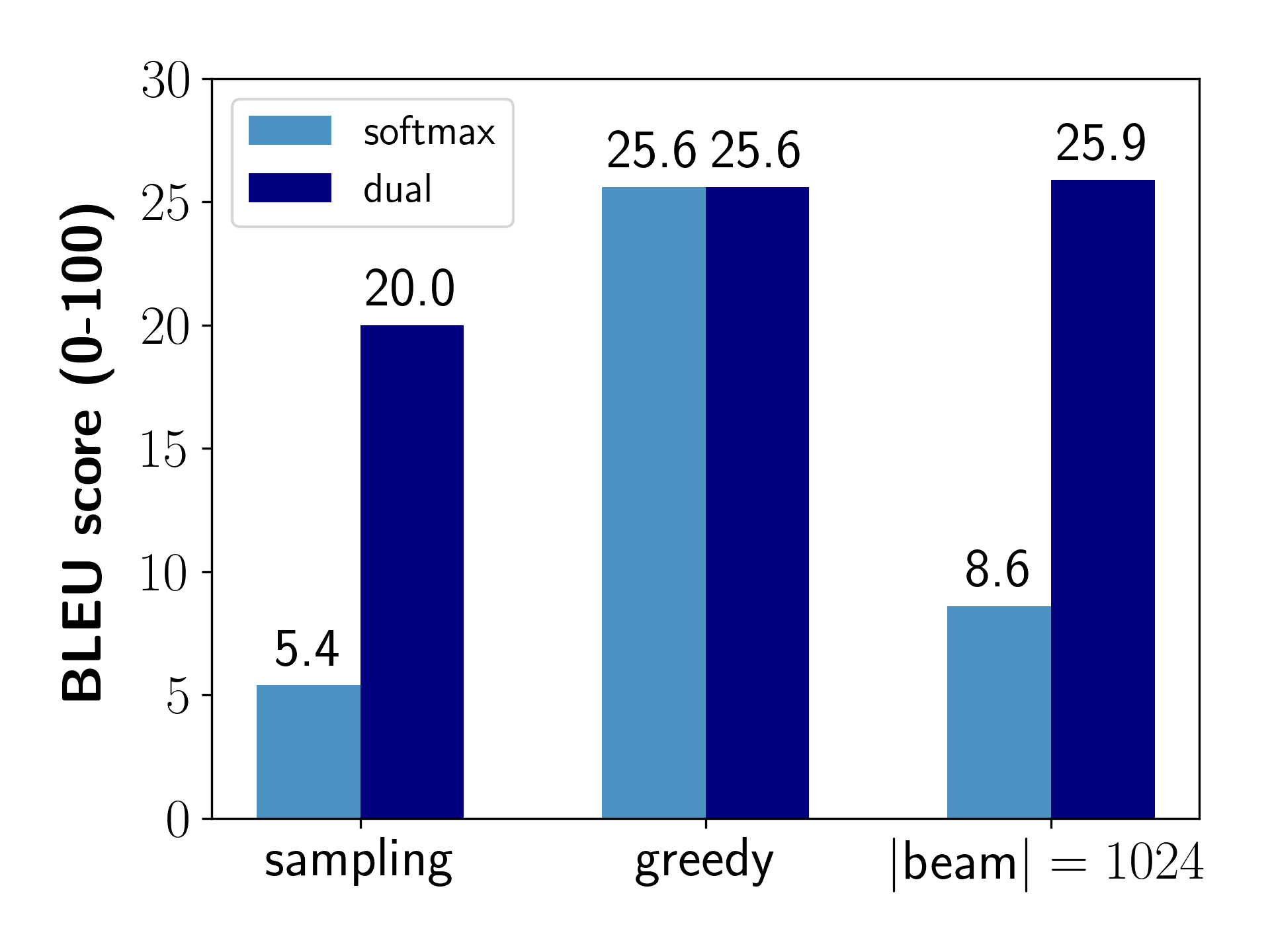}
\end{minipage}
\hfill
\begin{minipage}{0.4\linewidth}
\centering
\includegraphics[width=\linewidth]{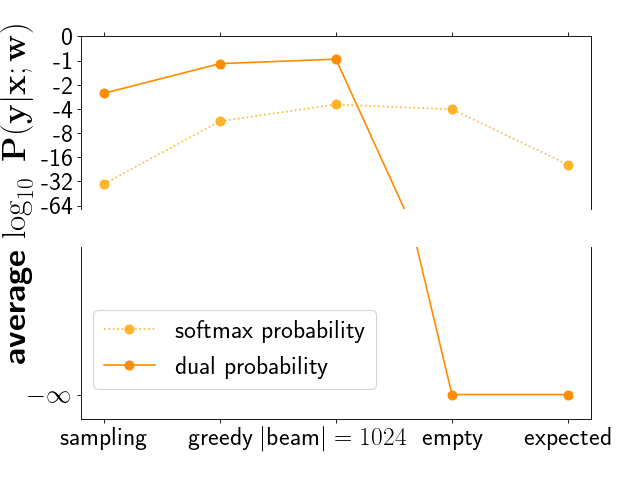}
\end{minipage}
\caption{On WMT'14 en$\ra$de, dual probabilities give much more reasonable probability predictions.}
\label{fig:wmt14_dual}
\end{figure}

\begin{figure}[H]
\begin{minipage}{0.45\linewidth}
\centering
\includegraphics[width=\linewidth]{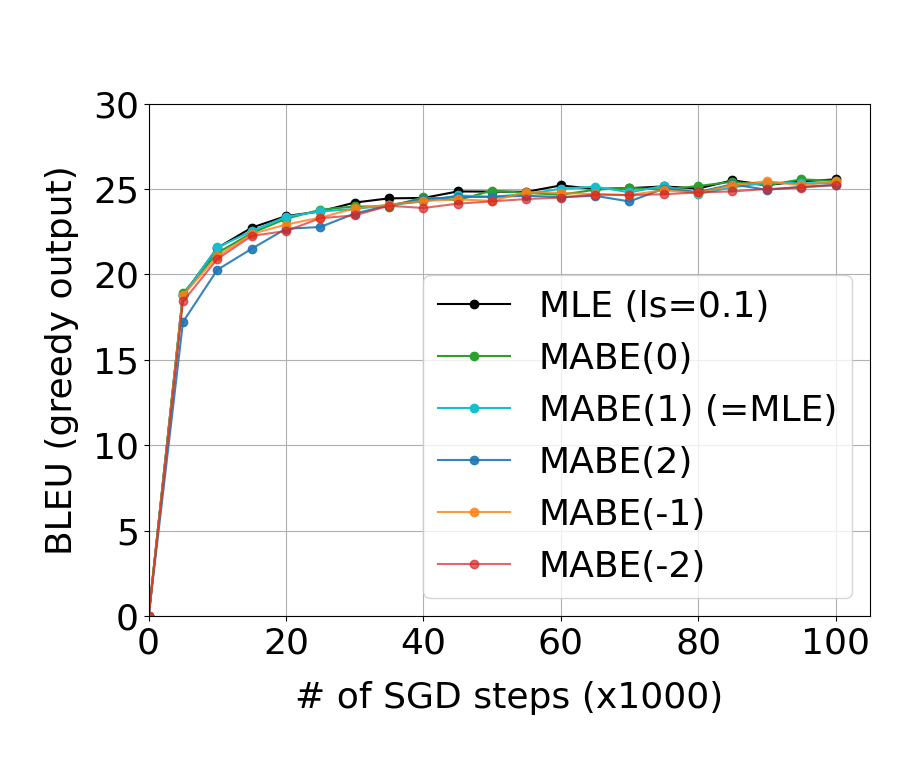}
\end{minipage}
\hfill
\begin{minipage}{0.5\linewidth}
\centering
\begin{tabular}{l|ccc}  
	\toprule
	& \multicolumn{3}{c}{BLEU (greedy)} \\
	& 	@90k steps  &	@95k steps  & @100k steps \\
	\midrule
	MLE (ls=0.1)		& 25.2			& 25.5			& 25.6 			\\
	\textbf{MABE(0)} 	& \textbf{25.2} & \textbf{25.6} & \textbf{25.5}	\\
	\textbf{MABE(1)} 	& \textbf{25.4} & \textbf{25.3}	& \textbf{25.3}	\\
	MABE(2) 			& 25.0 			& 25.1			& 25.2 			\\
	MABE(-1) 			& 25.5 			& 25.2			& 25.5			\\
	MABE(-2) 			& 25.0 			& 25.1			& 25.2			\\
	\bottomrule
\end{tabular}
\end{minipage}
\caption{On WMT'14 en$\ra$de, different MABE($\lambda$) variants have similar learning dynamics.}
\label{fig:wmt14_mabe}
\end{figure}

\newpage

\begin{figure}[t]
\begin{minipage}{0.5\linewidth}
\includegraphics[width=\linewidth]{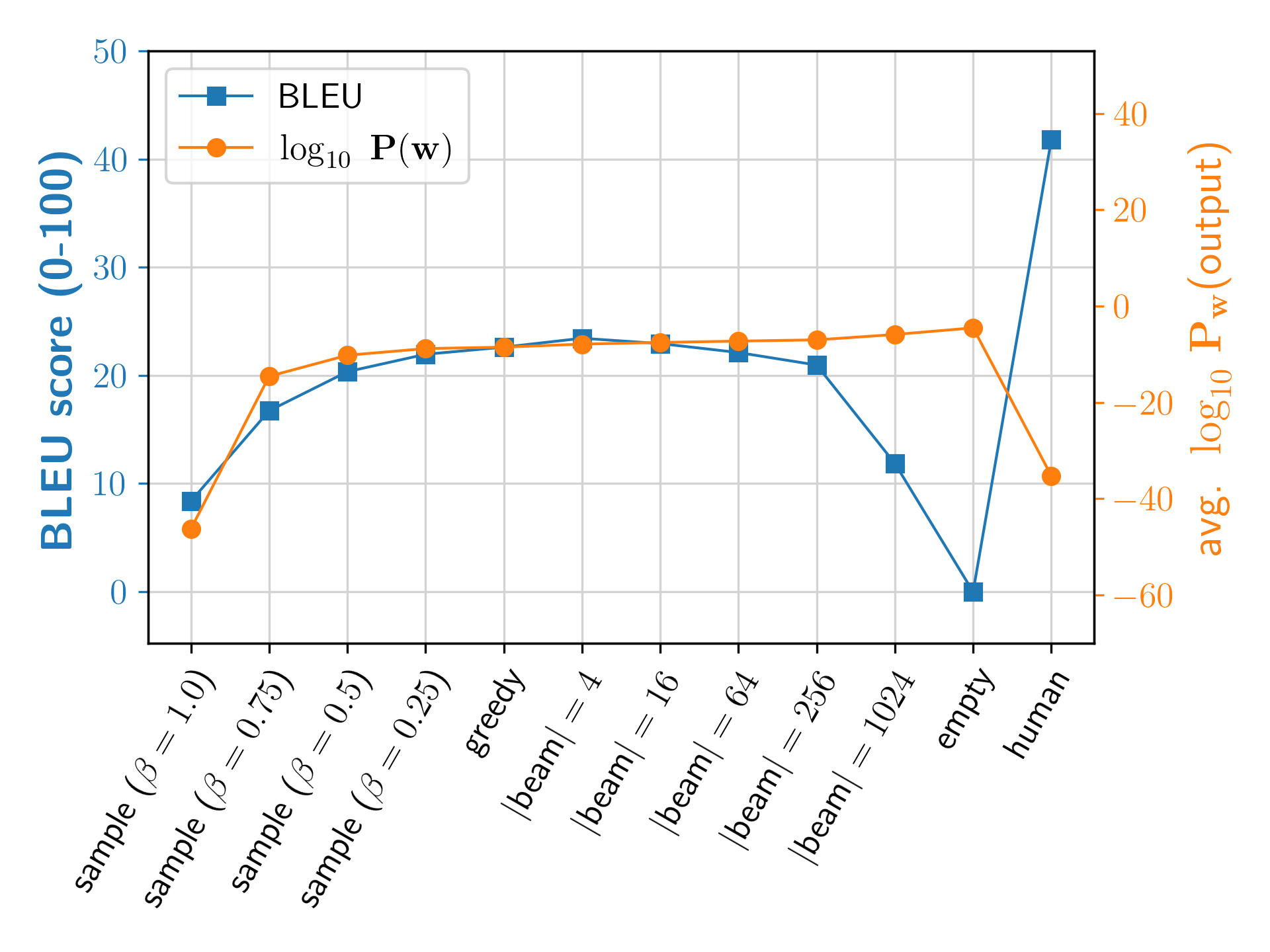}
\end{minipage}
\hfill
\begin{minipage}{0.4\linewidth}
\centering
\scalebox{0.95}{\begin{tabular}{|l|r|r|}
	\hline
	& BLEU 	& $\log_{10} \Pw$ \\
	\hline
	sampling 			& 8.3	& -46.3 \\
	\hline
	$\beta=0.75$ 		& 16.7 & -14.6 \\
	$\beta=0.5$ 		& 20.3 & -10.2 \\
	$\beta=0.25$ 		& 22.0 & -8.8 \\
	\hline
	greedy 				& 22.6 & -8.5 \\
	\hline
	$|\text{beam}|=4$ 	& 23.4 & -7.9 \\
	$|\text{beam}|=16$ 	& 22.9 & -7.5 \\
	$|\text{beam}|=64$ 	& 22.1 & -7.3 \\
	$|\text{beam}|=128$ & 21.7 & -7.1 \\
	$|\text{beam}|=256$	& 21.0 & -7.0 \\
	$|\text{beam}|=512$	& 19.3 & -6.8  \\
	$|\text{beam}|=1024$ & 11.9 & -5.9 \\
	\hline
	empty 				& 0.00 	& -4.5 \\
	\hline
	expected 			& 41.82~\footnote{See calculation method in Section \ref{sec:ref_perf}.} & -35.2 \\
	\hline
\end{tabular}}
\end{minipage}
\caption{On WMT'17 zh$\ra$en, MLE model exhibits paradoxical behaviors.}
\label{fig:wmt17_mle}
\end{figure}

\begin{figure}[H]
\begin{minipage}{0.4\linewidth}
\centering
\includegraphics[width=\linewidth]{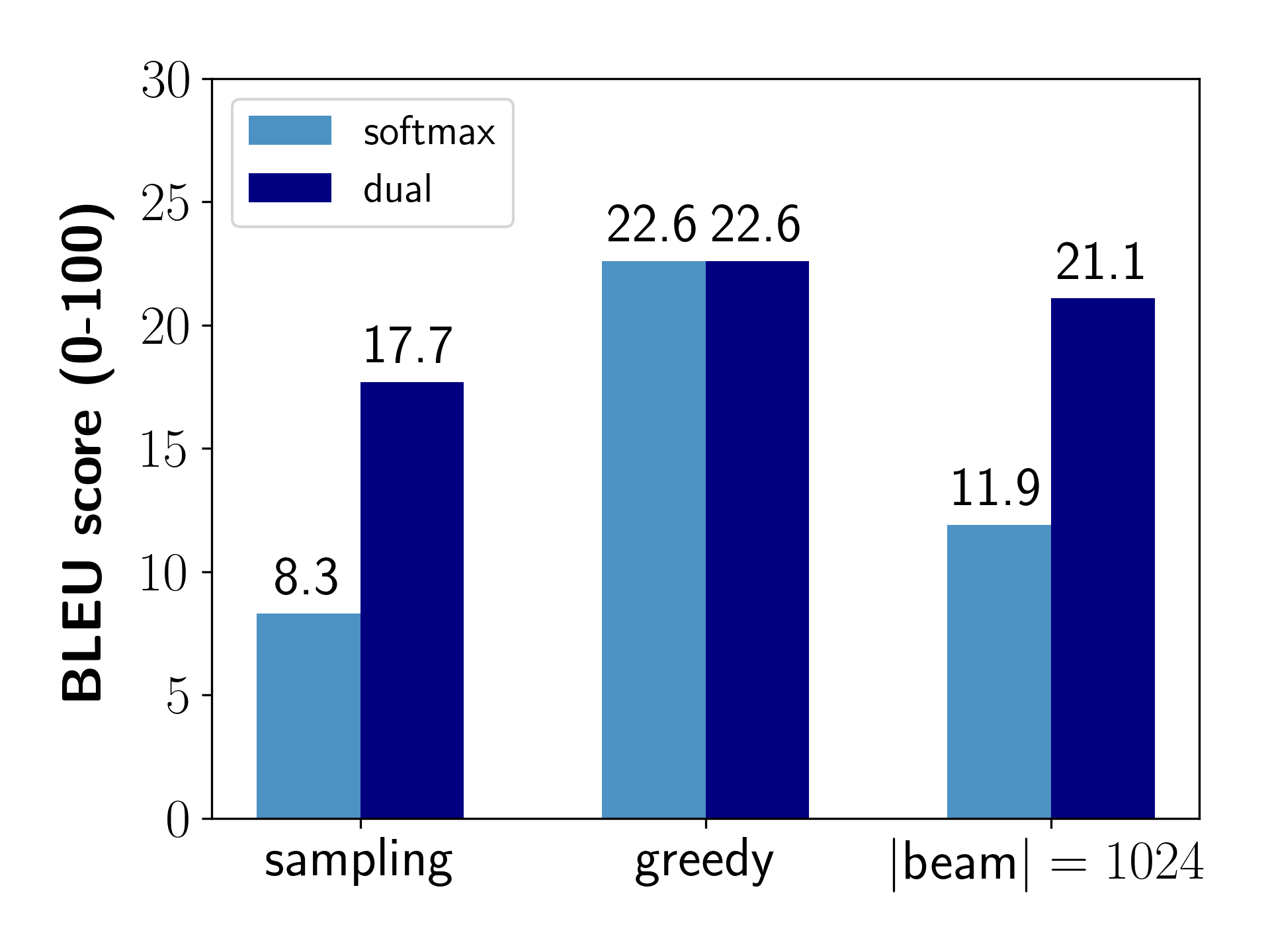}
\end{minipage}
\hfill
\begin{minipage}{0.4\linewidth}
\centering
\includegraphics[width=\linewidth]{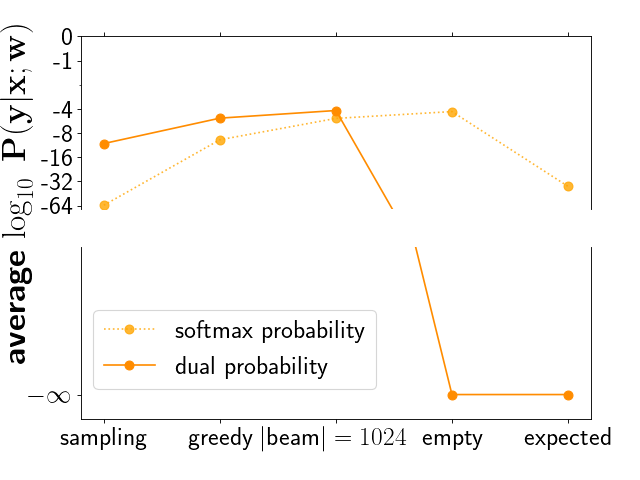}
\end{minipage}
\caption{On WMT'17 zh$\ra$en, dual probabilities give much more reasonable probability predictions.}
\label{fig:wmt17_dual}
\end{figure}

\begin{figure}[H]
\begin{minipage}{0.45\linewidth}
\centering
\includegraphics[width=\linewidth]{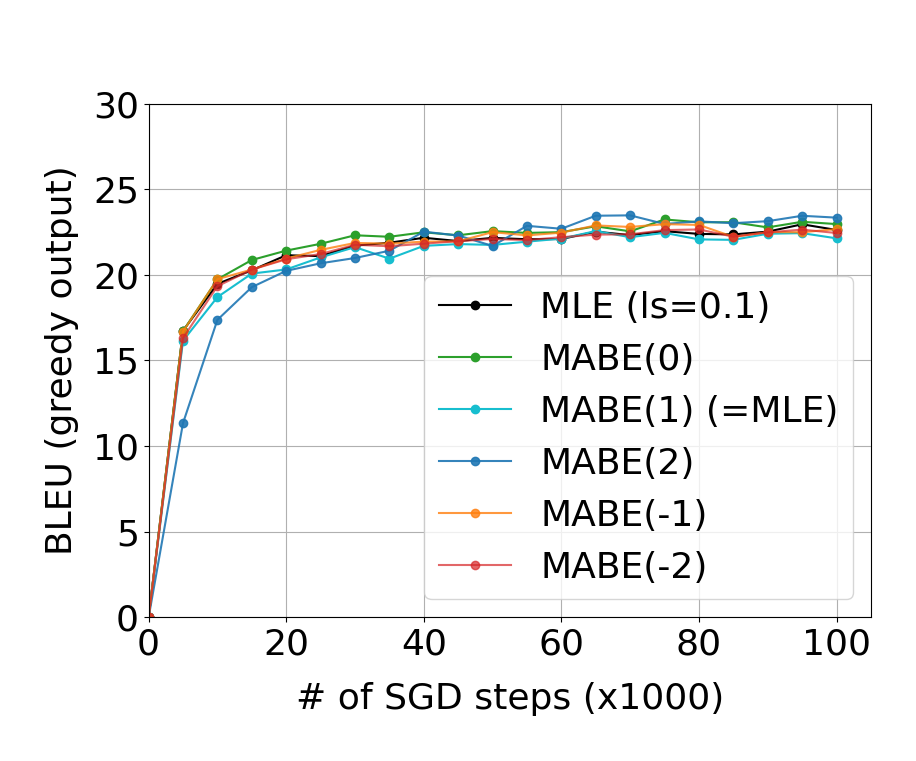}
\end{minipage}
\hfill
\begin{minipage}{0.5\linewidth}
\centering
\begin{tabular}{l|ccc}  
\toprule
& \multicolumn{3}{c}{BLEU (greedy)} \\
& 	@90k steps  &	@95k steps  & @100k steps \\
\midrule
MLE (ls=0.1)		& 22.5			& 22.9			& 22.6 			\\
\textbf{MABE(0)} 	& \textbf{22.8} & \textbf{23.1} & \textbf{23.0}	\\
\textbf{MABE(1)} 	& \textbf{22.4} & \textbf{22.4}	& \textbf{22.1}	\\
MABE(2) 			& 23.1 			& 23.4			& 23.3 			\\
MABE(-1) 			& 22.5 			& 22.5			& 22.7			\\
MABE(-2) 			& 22.5 			& 22.6			& 22.4			\\
\bottomrule
\end{tabular}
\end{minipage}
\caption{On WMT'17 zh$\ra$en, different MABE($\lambda$) variants have similar learning performance.}
\label{fig:wmt17_mabe}
\end{figure}

\newpage
\begin{figure}[t]
\begin{minipage}{0.5\linewidth}
\centering
\includegraphics[width=\linewidth]{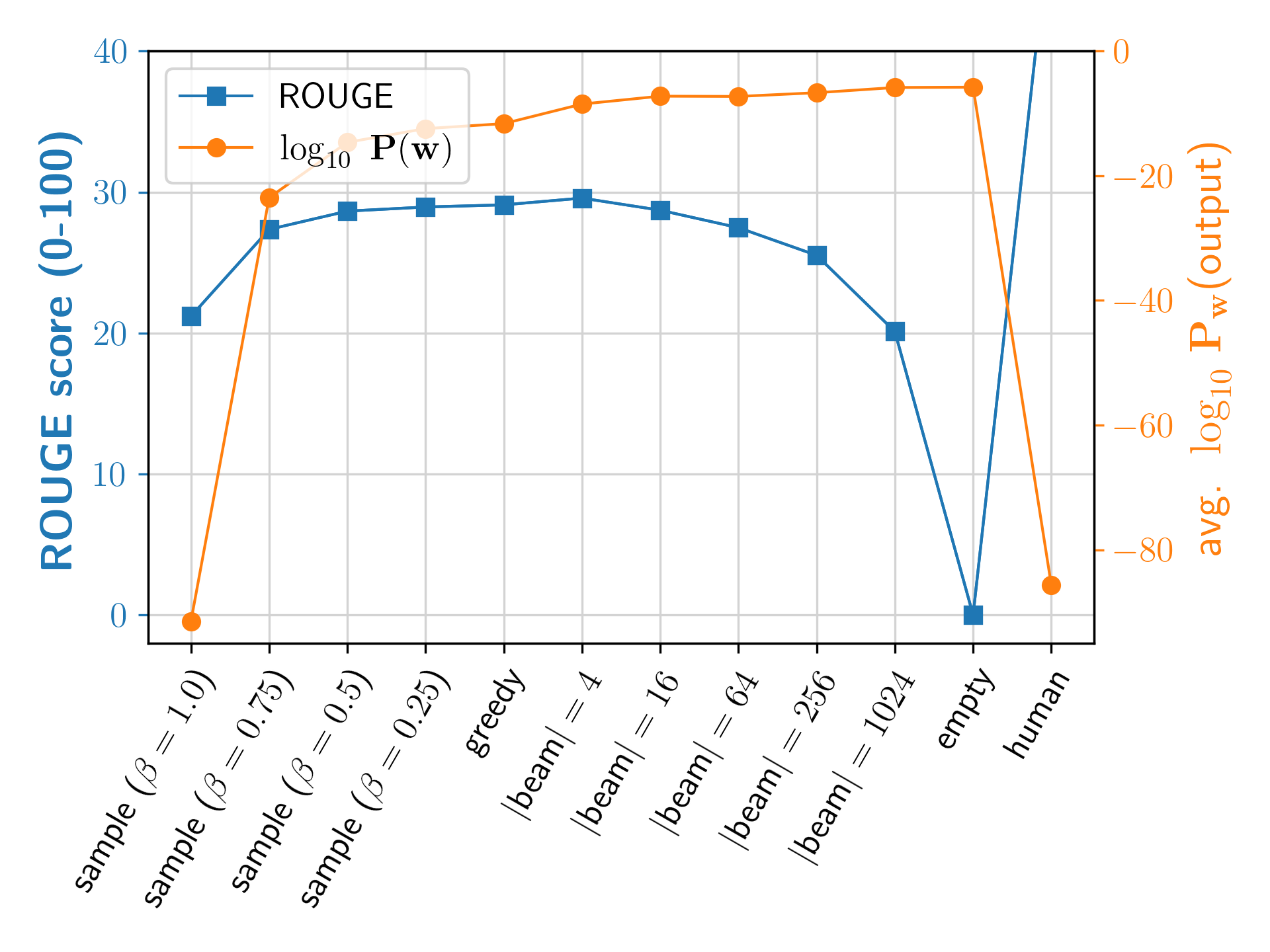}
\end{minipage}
\hfill
\begin{minipage}{0.4\linewidth}
\centering
\scalebox{0.95}{\begin{tabular}{|l|c|c|}
\hline
& ROUGE 	& $\log_{10} \Pw$ \\
\hline
sampling 	 			&	21.1 	&	-91.5 	\\ 
\hline
$\beta=0.75$ 			&	27.4 	&	-23.5 	\\
$\beta=0.5$	 			& 	28.6 	&	-14.6 	\\
$\beta=0.25$ 			&	28.9 	&	-12.4 	\\
\hline
greedy 					&	29.1 	&	-11.6 	\\
\hline
$|\text{beam}|=4$ 		&	29.6 	&	-8.5 	\\
$|\text{beam}|=16$ 		&	28.7 	&	-7.2 	\\
$|\text{beam}|=64$ 		&	27.5 	&	-7.3 	\\  
$|\text{beam}|=128$ 	&	26.1 	& 	-6.5	\\
$|\text{beam}|=256$ 	&	25.5 	&	-6.7 	\\  
$|\text{beam}|=1024$ 	&	20.1 	&	-5.9 	\\
\hline
empty 					&	0.00 	&	-5.8 	\\
\hline
expected &	100.00 ~\footnote{Lack of multi-reference data to calculate the actual ROUGE scores. ROUGE=100 for the reference summarizations in single-reference data set.} &	-85.7 \\
\hline
\end{tabular}}
\end{minipage}
\caption{On CNN/DailyMail, MLE model exhibits paradoxical behaviors.}
\label{fig:cnnmd_mle}
\end{figure}

\begin{figure}[H]
\begin{minipage}{0.4\linewidth}
\includegraphics[width=\linewidth]{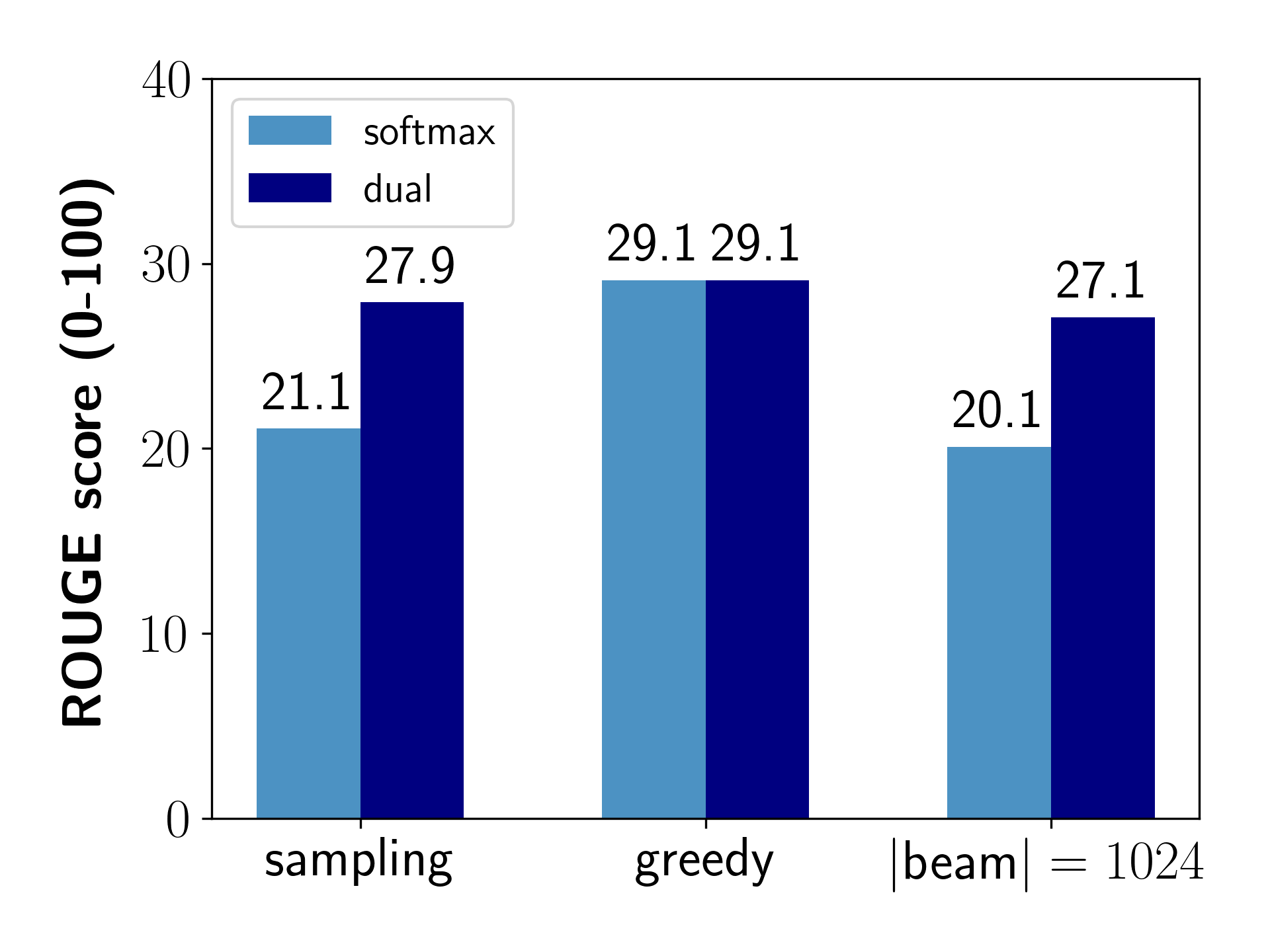}
\end{minipage}
\hfill
\begin{minipage}{0.4\linewidth}
\includegraphics[width=\linewidth]{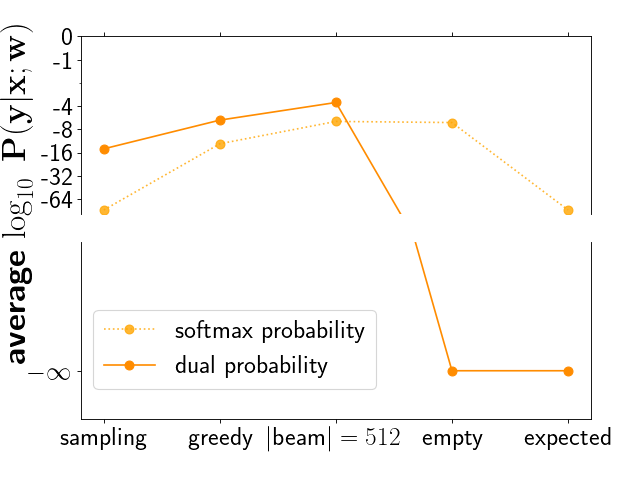}
\end{minipage}
\caption{On CNN/DailyMail, dual probabilities give more reasonable probability predictions.}
\label{fig:cnndm_dual}
\end{figure}

\begin{figure}[H]
\begin{minipage}{0.4\linewidth}
\centering
\includegraphics[width=\linewidth]{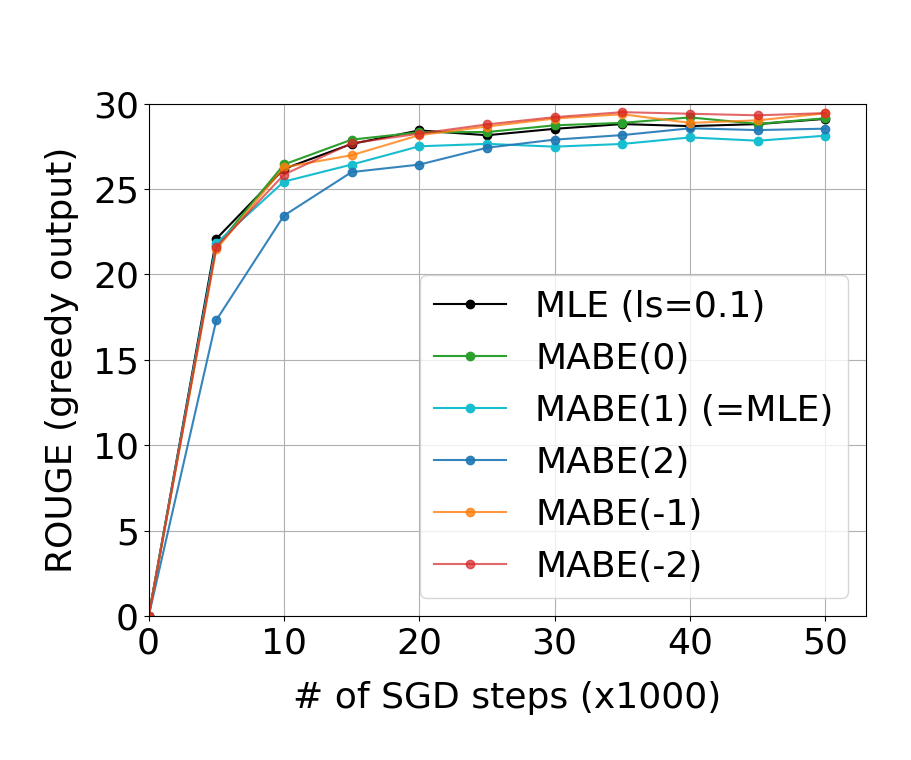} 
\end{minipage}
\hfill
\begin{minipage}{0.5\linewidth}
\centering
\begin{tabular}{l|ccc}  
\toprule
& \multicolumn{3}{c}{ROUGE (greedy)} \\
& 	@40k steps  &	@45k steps  & @50k steps \\
\midrule
MLE (ls=0.1)		& 28.7			& 28.8			& 29.1 			\\
\textbf{MABE(0)} 	& \textbf{29.2} & \textbf{28.8} & \textbf{29.1}	\\
\textbf{MABE(1)} 	& \textbf{28.0} & \textbf{27.8}	& \textbf{28.1}	\\
MABE(2) 			& 28.5 			& 28.4			& 28.5 			\\
MABE(-1) 			& 28.9 			& 29.0			& 29.4			\\
MABE(-2) 			& 29.4 			& 29.3			& 29.4			\\
\bottomrule
\end{tabular}
\end{minipage}
\caption{On CNN/DailyMail, different MABE($\lambda$) variants have similar learning performance.}
\label{fig:cnnmd_mabe}
\end{figure}

\newpage
\subsection{COMET Scores of the Dual Probability Model}
\label{sec:comet}

In above we have been using the standard performance metrics to evaluate probability models (BLEU~\cite{2002:bleu} for machine translation, and ROUGE~\cite{2004:rouge} for document summarization). These metrics use n-gram matching at the utterance level to evaluate similarity of two texts, and are under challenge when serving as performance metrics for tasks that require semantic-level paraphrase matching. To demonstrate that the performance gains by replacing softmax probabilities with dual probabilities are not a result of the flaw of these metrics themselves, we evaluate using a neural-network based metric called COMET~\cite{2020:comet}, which was recently proposed as an alternative to the traditional ``heuristic''-based metrics. As shown in Table \ref{tab:comet}, the trend is quite similar to what we observed in the previous experimentation sections, that the dual probability significantly boosts the COMET score of probability sampling and beam search with large beam, compared with the performance when using the softmax probability.

\begin{table}[t]
	\small
	\centering
	\caption{COMET scores of the dual probability model compared with the softmax probability model.}
	\scalebox{0.95}{\begin{tabular}{l|rr|rr|rr}  
			\toprule
			& \multicolumn{2}{|c|}{WMT'14 en$\ra$de} &
			\multicolumn{2}{c|}{WMT'17 zh$\ra$en}	&
			\multicolumn{2}{c}{CNN / DailyMail} \\
			& 	sampling &	$|\text{beam}|=1024$ &	sampling &	$|\text{beam}|=1024$ &	sampling &	$|\text{beam}|=1024$ \\
			\midrule
			Softmax Probability & 0.38 	& 0.53			& 0.59			& 0.64		& 0.45		& 0.52 \\
			Dual Probability 	& 0.67	& 0.72 			& 0.76			& 0.78 		& 0.60	 	& 0.61  \\
			& (+0.29)		& ~(+ 0.19) 		& ~(+ 0.17) 		& (+ 0.14) 		& (+ 0.15) 		& (+ 0.09) \\
			\midrule
			Greedy	 & \multicolumn{2}{|c|}{0.71} 	& \multicolumn{2}{|c|}{0.79} 	& \multicolumn{2}{|c|}{0.62}	   \\
			\bottomrule
		\end{tabular}
	}
	\label{tab:stat-dataset}
\end{table}

\subsection{Estimating The Performance of Expected Outputs}
\label{sec:ref_perf}

Accurately benchmarking human's performance is important for us to calibrate our understanding and expectation on the results from AI systems. To fairly evaluate the quality of human translation, it is important that we use \emph{independent} samples of the ``actual human translation'' $A$ and of the ``expected human translation'' $Y$ to calculate the BLEU scores. However, the standard WMT dataset only provide a single reference translation for each source sentence, in which case $A$ and $Y$ are strongly coupled and would always lead to a BLEU of strictly 100, which over-estimated the human performance. Fortunately, \citet{2018:multi-ref} released a dataset which provides ten human translations (per source sentence) for 500 instances in the WMT'14 en$\ra$de test set. See Table \ref{tab:simple-case} for an example of the multi-reference data. For WMT'17 zh$\ra$en, the CWMT2008 dataset\cite{2009:cwmt} provides a multi-reference data for Chinese$\ra$English news translation in which each source sentence is attached with four human translations. We used these multi-reference datasets to measure the task score of expected output (i.e. human translation) in Figure \ref{fig:wmt14_mle} and \ref{fig:wmt17_mle}. 

Specifically, for each source sentence $\x[i]$ in the multi-reference dataset, we randomly sampled a human translation as the expected output $\y[i]$, then randomly sampled (with replacement) another human translation as the actual output $\vect{a}\i$. The (independently) sampled expected and actual outputs for the whole corpus are then fed to the SacreBLEU script to compute a BLEU score. This process was repeated for 50 times to guarantee statistical significance. For English$\ra$German translation, the mean corpus-BLEU score of human translations is 42.54 (95\% confidence interval: 42.15 - 42.93) 
For Chinese$\ra$English translation, the mean corpus-BLEU score is 41.82 (95\% confidence interval: 41.43 - 42.20).

For the CNN/DailyMail summarization task, we did not find multi-reference dataset, thus simply marked 100 for human outputs. 

\begin{table}[t]
\caption{An example of multi-reference translations for WMT'14 en2de. Source is the source sentence \#1 in the dataset, Target is the official translation in \emph{newstest2014}, and Reference1-10 are the additional translations provided by \cite{2018:multi-ref}.}
\centering 
\footnotesize
\begin{tabular}{|p{12cm}|}
\hline
\textbf{[Source]:} Orlando Bloom and Miranda Kerr still love each other \\
\textbf{[Target]:}  Orlando Bloom und Miranda Kerr lieben sich noch immer\\ 
\textbf{[Reference1]:} Orlando Bloom und Miranda Kerr lieben sich noch\\
\textbf{[Reference2]:} Orlando Bloom und Miranda Kerr lieben sich immer noch .\\
\textbf{[Reference3]:} Orlando Bloom und Miranda Kerr lieben sich noch immer .\\
\textbf{[Reference4]:} Orlando Bloom und Miranda Kerr lieben sich immer noch .\\
\textbf{[Reference5]:} Orlando Bloom und Miranda Kerr lieben einander immer noch\\
\textbf{[Reference6]:} Orlando Bloom und Miranda Kerr lieben einander immer noch\\
\textbf{[Reference7]:} Orlando Bloom und Miranda Kerr lieben sich immer noch\\
\textbf{[Reference8]:} Orlando Bloom und Miranda Kerr lieben sich immer noch\\
\textbf{[Reference9]:} Orlando Bloom und Miranda Kerr lieben sich immer noch\\
\textbf{[Reference10]:} Orlando Bloom und Miranda Kerr lieben sich noch immer .\\
\hline
\end{tabular}

\label{tab:simple-case}
\end{table}


\vspace{1in}
\subsection{Reproducibility and Source Code}
\label{sec:code}

We provided our research source code in supplementary material to facilitate reproducibility. The README.md file gives detailed instructions to run
\squishlist
\item Experiment1 (Figure 2, 7, 10, 13): train a model with MABE($\lambda$) losses and with the label-smoothed MLE loss
\item Experiment2 (Figure 1, 5, 8, 11): run different decision rules based on the learned softmax probability model
\item Experiment3 (Table 1, Figure 6, 9, 12): run the same set of decision rules based on the learned dual probability model
\squishend
on all the three tasks as discussed above: WMT'14 en2de, WMT'17 zh2en, and CNN/DailyMail.

The entire experimentation pipeline is fully de-randomized once the random seed is specified. It takes roughly 240 hours to run the full experiment pipeline with one seed, on an A100 GPU. In total, the multi-seed experiment results presented in the paper take about 800 GPU hours (for A100). 

Finally, as an implementation trick to conveniently compute the covariance term \eqref{cov} with automatic differentiation library (e.g. pytorch), we utilized the equation \eqref{covt_proof2} in Section \ref{sec:proof_mle}, which gives
\begin{align}
&	\Cov[\Hpi_t \sim P_t(\w)]{Q_t(\Hpi_t;\w) ~,~ \nabla_{\w} Q_t(\Hpi_t;\w)} ~~\Big|_{\w=\w^+} \\
=~&	\sum_a P_t(a;\w) ~ Q_t(a;\w) ~ \Big(~ 
\nabla_{\w} Q_t(a;\w) - \sum_b P_t(b;\w) \nabla_{\w} Q_t(b;\w) 
~\Big) ~~\Big|_{\w=\w^+} \nonumber\\
=~&	\sum_a P_t(a;\w^+) ~ Q_t(a;\w^+) ~ \Big(~ 
\nabla_{\w} Q_t(a;\w) - \sum_b P_t(b;\w^+) \nabla_{\w} Q_t(b;\w) 
~\Big) ~~\Big|_{\w=\w^+} \nonumber\\
=~&	\nabla_{\w} \Bigg(~ \underline{
\sum_a P_t(a;\w^+) ~ Q_t(a;\w^+) ~ \Big(~ 
Q_t(a;\w) - \sum_b P_t(b;\w^+) Q_t(b;\w) 
~\Big) 
} ~\Bigg) ~~\Big|_{\w=\w^+} \label{covF_loss}\\
\end{align}
So, in our pytorch-based implementation, we first compute the function in \eqref{covF_loss} (the part highlighted by the underline), scale it by $\lambda$ and add to the $\Jmabe$ function, then perform back-propagation over the composite loss, which will give the MABE($\lambda$) gradient as prescribed in Algorithm \ref{algo:mabe}, due to \eqref{covF_loss}.

\section{Limitations and Future Works}
\label{sec:future_works}

In this section we discuss limitations of the current work, as well as the opportunities for future works.

First of all, the current paper focuses on studying a foundation of machine learning, and specifically, seeking to challenge and improve a widely-held \emph{mindset} on a widely-used training procedure in machine learning. Consequently, this paper is largely a theory paper (where the theory is defended not by pure mathematics but by a combination of mathematical \emph{and} experimental evidences). The primary goal here is thus not to propose new algorithms that immediately give empirical gains. 

However, we believe our utility-based theory implies many opportunities to invent such new algorithms in the future. One direction is to investigate deeper into the MABE($\lambda$) algorithm family. In our experiments, we see that the commonly used MABE(1) (a.k.a. MLE) is usually not the best-performing variant in this family, and the unperturbed variant MABE(0) often performs slightly better (e.g. compare the two in Figure \ref{fig:wmt14_mabe}, \ref{fig:wmt17_mabe}, and \ref{fig:cnnmd_mabe}). The MABE($\lambda$) algorithms presented in this paper are in the ``vanilla version'', without tricks such as label smoothing; the hyperparameters are set to the optimal setting for the MLE baseline (because our goal is not to beat it but to subsume it). It would be interesting works to develop and fine-tune the MABE($\lambda$) into a more fully-fledged solution.  

The dual probability formula \eqref{dual_prob} employed a simple heuristic way to ``adjust'' the probability predictions -- by first clipping to $[0,1]$ then normalizing the sum. As the choice here is somewhat arbitrary, it is possible that this adjustment could distort the probability predictions (with the payoff of preserving the axiom of probability). Note that such adjustment is only needed because the Q-values are not perfectly learned. Further investigation on the best way to refine the dual probability prediction here can be an interesting future work.

We also note that the beam search with beam size $=1024$ in our experiment would take much longer time than greedy decoding (50x for translation and 150x for summarization) -- it is clear that beam search is a bottleneck if we seriously want to explore the combinatorial solution space -- previously such exploration is not meaningful given the pathological behavior of traditional softmax probability. It would be interesting to see if more advanced search algorithms (such as Monte Carlo Tree Search) can better utilize the dual probability model.

Finally, there are also open questions in our value-based interpretation. For example, one limitation of the current theory is that it cannot explain why near-greedy outputs, such as search with a small beam size (e.g. 4) based on the ``problematic'' softmax probability can slightly outperform pure greedy outputs (although the margin is limited -- e.g. $27.1$ vs $25.6$ in WMT'14 en2de -- and the gain quickly disappears as the search scales up). Of course, it would be always interesting to test our theory in more machine learning tasks.

\end{document}